\documentclass[10pt,twocolumn,letterpaper]{article}

\usepackage{iccv}
\usepackage{times}
\usepackage{epsfig}
\usepackage{graphicx}
\usepackage{amsmath}
\usepackage{amssymb}
\usepackage{enumitem}
\usepackage{multirow}
\usepackage{float}
\usepackage{makecell}
\usepackage{verbatim}
\usepackage[section]{placeins}
\usepackage{mathtools}
\usepackage[normalem]{ulem}
\usepackage{color}
\usepackage{textcomp}

\usepackage[pagebackref=true,breaklinks=true,letterpaper=true,colorlinks,bookmarks=false]{hyperref}

\iccvfinalcopy 

\def\iccvPaperID{3710} 

\ificcvfinal\fi

\begin{document}

\title{Learning to Match Features with Seeded Graph Matching Network}

\author{Hongkai Chen$^{1}$\hspace{0.15cm} Zixin Luo$^{1}$\hspace{0.15cm} Jiahui Zhang$^{2}$\hspace{0.15cm} Lei Zhou$^{1}$\hspace{0.15cm} Xuyang Bai$^{1}$\hspace{0.15cm} Zeyu Hu$^{1}$\hspace{0.15cm}  \\Chiew-Lan Tai$^{1}$\hspace{0.15cm} Long Quan$^{1}$ \\
\normalsize $^1$Hong Kong University of Science and Technology \hspace{0.7cm} $^2$ Tsinghua University \hspace{0.7cm} \normalsize\\
\tt\small\{hchencf,zluoag,lzhouai,xbaiad,zhuam,taicl,quan\}@cse.ust.hk \hspace{0.2cm}
\tt\small jiahui-z15@mails.tsinghua.edu.cn
}

\maketitle
\ificcvfinal\thispagestyle{empty}\fi

\begin{abstract}
Matching local features across images is a fundamental problem in computer vision. Targeting towards high accuracy and efficiency, we propose Seeded Graph Matching Network, a graph neural network with sparse structure to reduce redundant connectivity and learn compact representation. The network consists of 1) Seeding Module, which initializes the matching by generating a small set of reliable matches as seeds. 2) Seeded Graph Neural Network, which utilizes seed matches to pass messages within/across images and predicts assignment costs. Three novel operations are proposed as basic elements for message passing: 1) Attentional Pooling, which aggregates keypoint features within the image to seed matches. 2) Seed Filtering, which enhances seed features and exchanges messages across images. 3) Attentional Unpooling, which propagates seed features back to original keypoints. Experiments show that our method reduces computational and memory complexity significantly compared with typical attention-based networks while competitive or higher performance is achieved. \vspace{-1.5em}
\end{abstract}

\let\saveFloatBarrier\FloatBarrier
\let\FloatBarrier\relax
\section{Introduction}
\let\FloatBarrier\saveFloatBarrier
Establishing reliable correspondences across images is an essential step to recover relative camera pose and scene structure in many computer vision tasks, sush as Structure-from-Motion (SfM)~\cite{sfm}, Multiview Stereo(MVS)~\cite{mvs} and Simultaneous Localization and Mapping (SLAM)~\cite{slam}. In classical pipelines, correspondences are obtained by nearest neighbour search (\textit{NN}) of local feature descriptors and are usually further pruned by heuristic tricks, such as mutual nearest neighbour check (\textit{MNN}) and ratio test (\textit{RT})~\cite{lowe2004distinctive}.

In the past few years, great efforts have been made on designing learnable matching strategy. Early works~\cite{pointcn,acnet,oanet} in this direction utilize PointNet~\cite{pointnet}-like networks to reject outliers of putative correspondences. In these works, the correspondence coordinates are fed into permutation-equivariant networks, then inlier likelihood scores are predicted for each correspondence. Despite showing exciting results, these methods are limited in two aspects: 1) They operate on pre-matched correspondences, whereas finding more matches than vanilla nearest-neighbour matching is impossible. 2) They only reason about the geometric distribution of putative correspondences, neglecting the critical information of original local visual descriptors.

\begin{figure}[t]
	\centering 
	\includegraphics[width=0.48\textwidth]{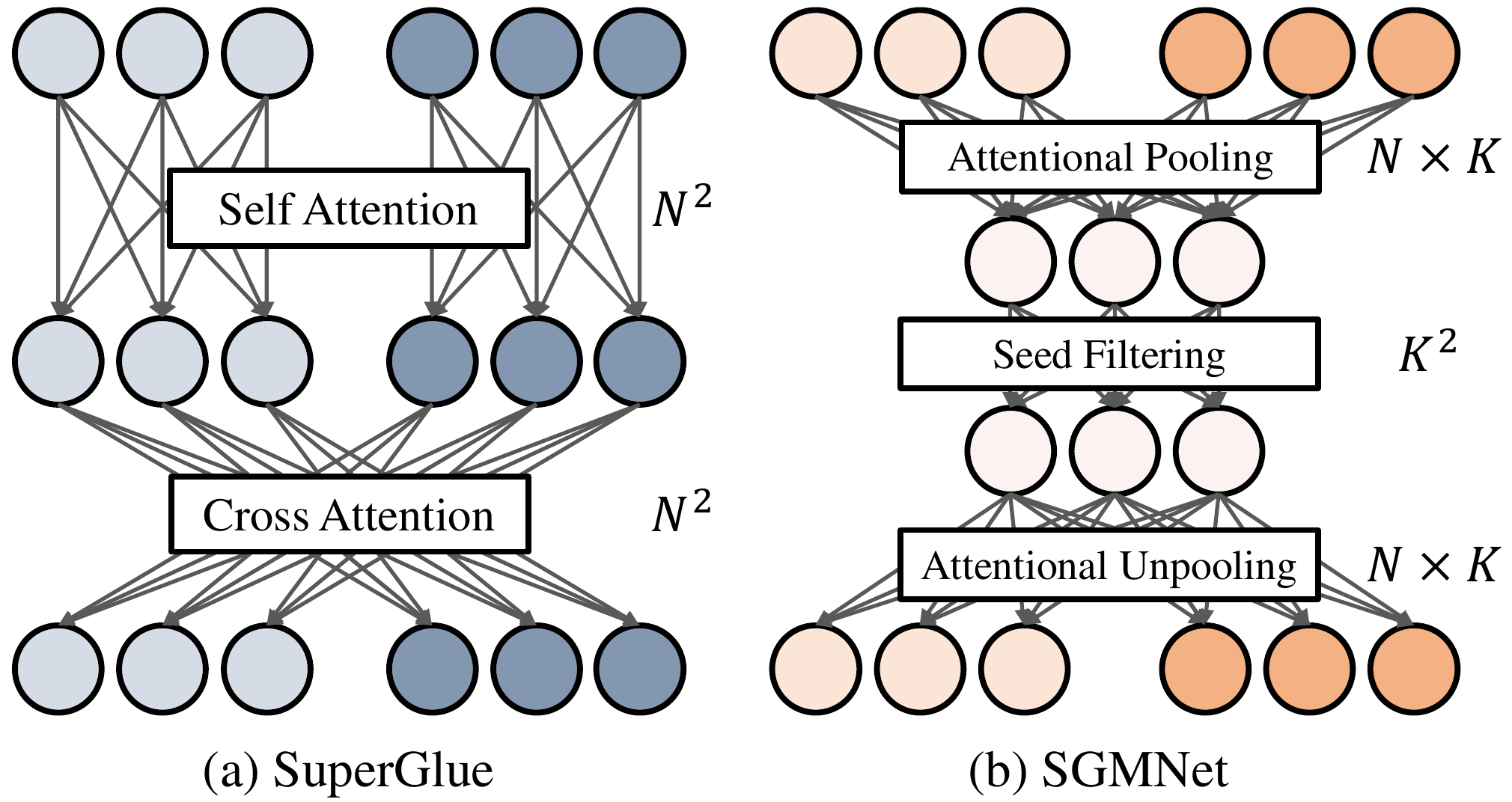}
	\caption{The designs of message passing layer. (a) SuperGlue densely connects every node in the graph, resulting in $\mathcal{O}(N^2)$ computational complexity. (b) Instead, the proposed network, SGMNet, adopts pooling/unpooling operations, reducing the complexity to $\mathcal{O}(NK)$, where $K\ll N$ in general.}
	\vspace{-2em}
	\label{fig:teaser}
\end{figure}

Another thread of methods cast feature matching as a graph matching problem~\cite{superglue,learngraph,handgraph}, which mitigates the limit of vanilla nearest neighbour correspondences. 
The representative work, SuperGlue~\cite{superglue}, constructs densely-connected graphs between image keypoints to exchange messages about both visual and geometric context. However, the superior performance comes along with high computation and memory cost, especially when being applied to keypoints of larger number (e.g. up to $10k$). As illustrated in Fig.~\ref{fig:teaser}(a), the message passing layer of SuperGlue first calculates similarity scores exhaustively between every two nodes, then gathers features to pass messages densely in the graph. This results in computational complexity of $\mathcal{O}(N^2C)$ for matrix multiplications, and memory occupation of $\mathcal{O}(N^2)$ to hold the attention matrix, supposing that the keypoint number is $N$ and feature channel is $C$. The complexity increases even drastically for deeper graph networks. Given this, exploring more efficient and compact message passing operation is of practical significance.

Besides the major efficiency bottleneck, it is debatable if such densely-connected graph introduces too much redundancy or insignificant message exchange that may hinder the representation ability, especially in the context of feature matching where the match set is highly outlier-contaminated and a large portion of keypoints are unrepeatable. As a result, most graph edges from SuperGlue~\cite{superglue} tend to have zero strength, as reported in its original paper and also observed in our experiments. This phenomenon indicates that even a  sparse graph is largely sufficient and less distracted from unnecessary message exchanges.

In this paper, we propose Seeded Graph Matching Network (SGMNet) to mitigate above limitations from two aspects. First, inspired by guided matching approaches~\cite{darmon2020learning,guidematching,statisticalflow}, we design a \textit{Seeding Module} that initializes the matching from a small set of reliable matches so as to more effectively identify inlier compatibility. Second, we draw inspiration from graph pooling operations~\cite{diffpool,pool2}, and construct a \textit{Seeded Graph Neural Network} whose  graph structure is largely sparsified to lower the computation and reduce the redundancy. Specifically, three operations are proposed to construct our message passing blocks. As illustrated in Fig.~\ref{fig:teaser}(b), instead of densely attending to all features within/across images, original keypoint features are first pooled by 1) \textit{Attentional Pooling} through a small set of seed nodes, of which the features will be further enhanced by 2) \textit{Seed Filtering}, and finally recovered back to original keypoints through 3) \textit{Attentional Unpooling}.

By using seeds as attention bottleneck between images, the computational complexity for attention is reduced from $\mathcal{O}(N^2C)$ to $\mathcal{O}(NKC)$, where $K$ is the number of seeds. When $K\ll N$, for example, $8k$ features are pooled into $512$ seeds, the actual computation will be significantly cut down. We evaluate SGMNet under different tasks to demonstrate both its efficiency and effectiveness, and summarize our contributions threefold:
\begin{itemize}[leftmargin=*]\itemsep0em
	\item A seeding mechanism is introduced in graph matching framework to effectively identify inlier compatibility.
	\item A greatly sparsified graph neural network is designed that enables more efficient and clean message passing.
	\item Competitive or higher accuracy is reported with remarkably improved efficiency over dense attentional GNN. As an example, when matching $10k$ features, SGMNet runs $7$ times faster and consumes 50\% less GPU memory than SuperGlue.
\end{itemize}

\vspace{-0.5em}
\section{Related Works}
\vspace{-0.8em}
\smallskip\noindent\textbf{Learnable image matching.} 
Integrating deep learning techniques into geometry-based computer vision task, such as MVS~\cite{mvsnet,rmvs,jy1,jy2} and Visual Localization~\cite{kfnet,shitao}, has received inspiring success during the past few years. As the front-end component for geometry estimation, learnable image matching has also been proven effective, where works in this area can be roughly divided into two categories. The first one focuses on improving local descriptors~\cite{lift,deepdesc,geodesc,contextdesc,mishchuk2017working,tian2017l2} and keypoints~\cite{tilde,d2net,asl,revaud2019r2d2,sp} with convolutional neural networks, while methods in the second category attempt to embed learning techniques into matching strategy, which involves learnable outlier rejection~\cite{pointcn,acnet,oanet} and robust estimator~\cite{ngransac}.

Recently, a new framework, SuperGlue~\cite{superglue}, is proposed to integrate feature matching and outlier rejection into a single graph neural network (GNN). Though exhibiting promising results in different tasks, SuperGlue still suffers from the excessive computational cost of fully connected self/cross attention operation, especially when used to match features in high number.

Compared with SuperGlue, our method shares the same advantages, that is, feature matching and refinement are integrated into one single network and allows for end-to-end training. However, our network significantly reduces the computational and memory cost due to its efficient attention block, which is specially designed for image matching.

\begin{figure*}[t]
	\centering
	\includegraphics[width=0.85\textwidth]{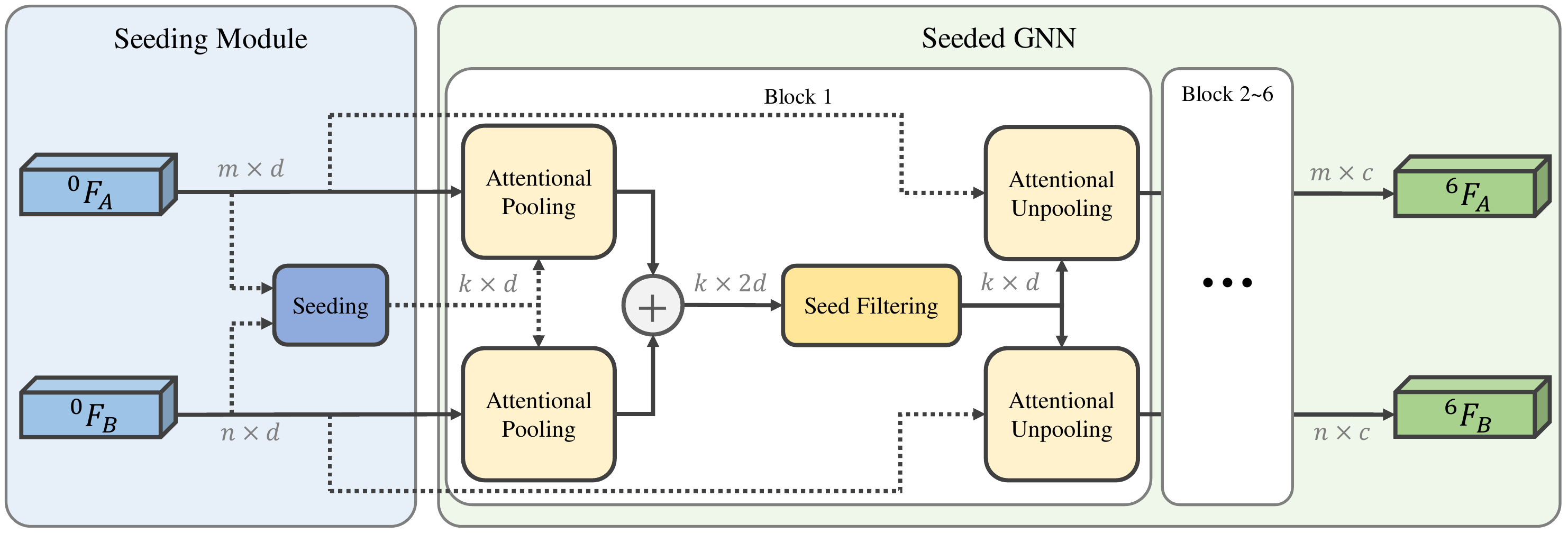}
	\caption{The network architecture of SGMNet, which takes the local features as input, then generates seed matches from \textit{seeding module}, and finally extracts correspondence features from \textit{Seeded GNN} with multiple attentional blocks. In practice, updated features will be fed into a reseeding module and a 3-layer seeded GNN for refinement, while we omit this procedure here for simplicity.
	}
	
	\label{fig:net_arch}
\end{figure*}

\smallskip\noindent\textbf{Efficient transformer architectures.} Transformer~\cite{transformer} architectures have gained intensive interest during the past few years. Specially, in the contexts of graph convolution, the attention mechanism in transformer can be used to pass messages across nodes in graph structure~\cite{deepgraph1,deepgraph2}. Despite its effectiveness in a wide range of tasks, one major concern about transformer is its quadratic complexity w.r.t. the input size, which hinders its application under large query/key elements number. 

Recently, many efforts have been made to address the attention efficiency problem. In~\cite{transpre1,transpre2} predefined sparse attention pattern are adopted to cut down memory/computation cost. In~\cite{transseg1,transseg2}, attention span is pruned by using learnable partitioning or grouping on input elements. In~\cite{transpool1,settrans}, pooling operation is utilized to reduce elements number. Despite the inspiring progress, works in this area generally focus on self-attention, where keys and queries are derived from the same element set, while their effectiveness in cross-attention, where keys and queries come from two unaligned sets, remains unstudied.

We draw inspiration from induced set attention (ISA)~\cite{settrans}, where a set of learned but fixed nodes are utilized as bottleneck for efficient self attention. To be compatible with cross attention in graph matching, we establish attention between seed matches and original point sets. The selected reliable correspondences align features in both sides and pass message in a low cost way.     

\smallskip\noindent\textbf{Graph matching.} Graph matching, which aims to generating node correspondences across graphs, is a widely used model for feature matching in both 2D~\cite{handgraph,learngraph} and 3D~\cite{sm,xy} domain. Mathematically formulated as a quadratic assignment problem (QAP)~\cite{gm}, graph matching is NP-hard in its most general form and requires infeasible expensive solver for precise solutions. Despite the intractable nature of general graph matching, some methods~\cite{sgm,sgm1,sgm2} leverage partially pre-matched correspondences, also called seeds, to help matching, which are referred to as Seeded Graph Matching(SGM). Inspired by SGM, our network integrate seeds into a GNN framework for compact message passing and robust matching.

\section{Methodology}
We present Seeded Graph Matching Network, shorten as SGMNet, for learning correspondences between two sets of keypoints and their associated visual descriptors. As illustrated in Fig.~\ref{fig:net_arch}, our network produces matches in two stages: 1) \textit{Seeding Module} generates seeds to guide compact message passing, and 2) \textit{Seeded Graph Neural Network} leverages seeds as message bottleneck to update per-node feature. In the following parts, an overview of our network architecture will be introduced first, followed by detailed description of each module.
\subsection{Overview}
Given a pair of images $\mathbf{A}$ and $\mathbf{B}$, with $n$ and $m$ keypoints and associated visual descriptors respectively, indexed by $\mathbf{\alpha}:=\{1,...,n\}, \mathbf{\beta}:=\{1,...,m\} $, our objective is to establish reliable and robust keypoint matches across two images. 

We formulate the keypoint matching task as a graph matching problem, where the nodes are the keypoints of each image.
Instead of apply fully connected graph, we generate a set of keypoint correspondences, which we refer to as \textit{seed matches}, to guide message pass across nodes in two graphs for subsequent matching. This critical difference allows for processing large number of keypoints with significantly lower memory and computation cost.

The input to our network are keypoints $\mathbf{K_A}$ , $\mathbf{K_B}$ in two images, $
\mathbf{K}_\mathbf{I}^i=(\mathbf{p}^i_\mathbf{I},\mathbf{d}^i_\mathbf{I})$, where $\mathbf{I} \in \mathbf{\{A,B\}}$
and $\mathbf{p}^i_\mathbf{I}=(x^i_\mathbf{I},y^i_\mathbf{I})$ is the coordinate of keypoint $i$ in image $\mathbf{I}$. $\mathbf{d}^i_\mathbf{I} \in  \mathbb R^d $ is associated $d$ dimensional visual descriptor.  

Positions of keypoints are embedded into high dimensional feature space and combined with descriptors by element-wise summation for initial representation $\mathbf{\prescript{0}{}{F_A}}$ , $\mathbf{\prescript{0}{}{F_B}}$.
\vspace{-1em}
\begin{align}
\mathbf{\prescript{0}{}{F}}^i_\mathbf{I}=&\mathbf{d}^i_\mathbf{I}+\text{MLP}(\mathbf{p}^i_\mathbf{I}), \mathbf{I} \in \mathbf{\{A,B\}},
\end{align}

A \textit{Seeding Module} follows to construct a set of seed matches $\mathbf{S}$. $\mathbf{\prescript{0}{}{F_A}}$, $\mathbf{\prescript{0}{}{F_B}}$ and $\mathbf{S}$ are then fed into our \textit{Seeded Graph Neural Network} which reasons about visual appearance similarity, neighbourhood consensus as well as the guidance provided by seed matches jointly to update keypoint features.

Inspired by the cascade refinement structure in OANet~\cite{oanet}, a second seeding module, or reseeding module, is introduced to generate more accurate seeds based on the updated features, which helps to further refine matches with another Seeded GNN. Final matches are then generated by formulating assignment matrix.

\subsection{Seeding Module}
Proposing a set of seed matches lays the foundation for subsequent matching. For initial seeding, we adopt a simple yet effective strategy: we generate putative matches by nearest neighbour matching and use the inverse of distance ratio, i.e., the ratio of distance to first and second nearest neighbours~\cite{lowe2004distinctive}, as reliability scores. We adopt Non-Maximum Suppression (NMS) for a better spatial coverage of seeds. More details of seeding module can be found in Appendix A.2. Despite potential noise in initial seeds, our network maintains robustness with proposed weighted unpooling operation and reseeding strategy, which will be discussed later.

The seeding module outputs $\mathbf{S}=\mathbf{(S_A,S_B)}$,
where $\mathbf{S_A}$, $\mathbf{S_B}$ are index lists for seed matches in each image.

\subsection{Seeded Graph Neural Network (Seeded GNN)}
Seeded GNN takes initial position-embedded features $\mathbf{\prescript{0}{}{F_A}}$, $\mathbf{\prescript{0}{}{F_B}}$ and leverages seed matches $\mathbf{S}$ as attention bottlenecks for message passing. To this end, we adopt a pooling-processing-unpooling strategy in each processing unit: seed features first gather information from full point sets on each side through \textbf{Attentional Pooling}, then processed by \textbf{Seed Filtering} operation and finally be recoverd back to original size by \textbf{Attentional Unpooling}. Our Seeded GNN is constructed by stacking 6(3) such processing units for initial(refinement) stages. 

\smallskip\noindent\textbf{Weighted attentional aggregation.} 
We first  introduce  a weighted  version of  attentional  aggregation, which allows for sharper and cleaner data-dependent message passing.

In a $d$-dimensional feature space, for $m$ vectors to be updated: $\mathbf{X}\in \mathbb R^{m \times d}$, $n$ vectors to be attended to: $\mathbf{Y}\in \mathbb R^{n \times d}$ and a weight vector: $\mathbf{w} \in \mathbb R^{n}$, the weighted attentional aggregation \textbf{Att} is defined as,

\begin{align}
\mathbf{X^r}=\textbf{\text{Att}}\mathbf{(X,Y,w)}=\mathbf{X+\text{MLP}(X||\Delta)},
\end{align}
where
\begin{align}
\Delta=\theta(\mathbf{QK}^T)\mathbf{WV}, \mathbf{W}=Diag(\mathbf{w}), \mathbf{W} \in \mathbb R^{n \times n}
\end{align}
$\theta(\cdot)$ means row-wise softmax. $\mathbf{Q}$ is linear projection of $\mathbf{X}$ and $\mathbf{K,V}$ are linear projections of $\mathbf{Y}$. $\mathbf{X^r}$ is renewed representation for $\mathbf{X}$.

By attentional aggregation, elements in $\mathbf{X}$ retrieve and aggregate information from elements in $\mathbf{Y}$. A weighting vector $\mathbf{w}$ is applied on $\mathbf{V}$ to adjust the importance for each element in $\mathbf{Y}$.

\smallskip\noindent\textbf{Attentional pooling.} As the first step in message passing, seed matches retrieves contexts from full keypoint set through attentional aggregration,

For input features $\mathbf{\prescript{t}{}{F_A}}$ , $\mathbf{\prescript{t}{}{F_B}}$ in layer $t$, features for seed matches are first retrieved by indices $\mathbf{(S_A,S_B)}$
\begin{align}
\mathbf{\prescript{t}{}{S}}_\mathbf{I}^1=\mathbf{\prescript{t}{}{F_I}[S_I]}, \mathbf{I \in \{A,B\}},
\end{align}
where $[~]$ is the indexing operation. Seed matches are then updated by retrieving context from nodes in each graph
\begin{align}
\mathbf{\prescript{t}{}{S}}_\mathbf{I}^2=\textbf{Att}(\mathbf{\prescript{t}{}{S}}_\mathbf{I}^1,\mathbf{\prescript{t}{}{F}}_\mathbf{I},\mathbf{1}),\mathbf{\ I \in \{A,B\}},
\end{align}
where $\mathbf{1}$ is all one vector, which means no weights are applied.

A mutilayer perceptron follows to fuse the seed features, 
\begin{align}
[\mathbf{\prescript{t}{}{S}}_\mathbf{A}^3||\mathbf{\prescript{t}{}{S}}_\mathbf{B}^3]=&\text{MLP}(\mathbf{\prescript{t}{}{S}}_\mathbf{A}^2||\mathbf{\prescript{t}{}{S}}_\mathbf{B}^2)
\end{align}
where $||$ means concatenation along row dimension.

The outputs $\mathbf{\prescript{t}{}{S}}_\mathbf{A}^3$, $\mathbf{\prescript{t}{}{S}}_\mathbf{B}^3$, which encode both visual and position context for each graph and information from seed matches themselves, are fed into subsequent operations.

\begin{figure}[t]
	\centering 
	\includegraphics[width=0.45\textwidth]{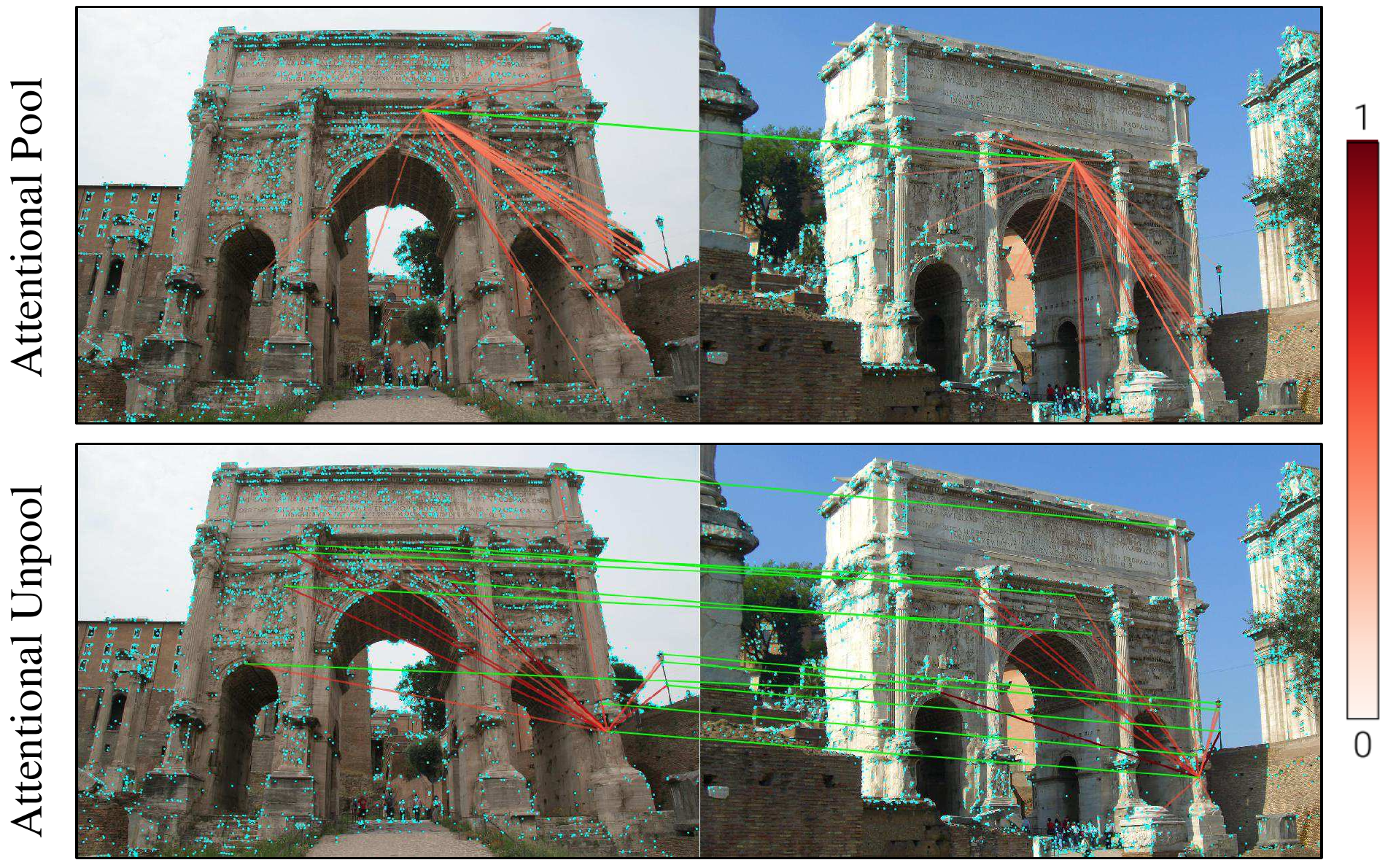}
	\caption{Visualization of attentional pooling/unpooling. In attentional pooling, one pair of seed match aggregates context from the other keypoints, while in attentional unpooling each original keypoint retrieves renewed message from seed matches. }
	\label{fig:att_vis}
	\vspace{-1em}
\end{figure}

\smallskip\noindent\textbf{Seed filtering.}
 We propose seed filtering operation to (1) conduct intra/inter-graph communication between seed matches and (2) suppress the influence of outlier seed matches. More specifically, intra/inter-graph attentional aggregation is applied to the input seed correspondence features  $\mathbf{\prescript{t}{}{S}}_\mathbf{I}^3 \in \mathbb R^{k \times d},\ \mathbf{I \in \{A,B\}}$.
\begin{align}
\mathbf{\prescript{t}{}{S}}_\mathbf{I}^4 &=\textbf{Att}(\mathbf{\prescript{t}{}{S}}_{\mathbf{I}}^3,\mathbf{\prescript{t}{}{S}}_\mathbf{I}^3, \mathbf{1}),\\
\mathbf{\prescript{t}{}{S}}_\mathbf{I}^5=\textbf{Att}(\mathbf{\prescript{t}{}{S}}_\mathbf{I}^4,&\mathbf{\prescript{t}{}{S}}_\mathbf{J}^4,\mathbf{1}),\ \mathbf{I,J \in \{A,B\}, I \neq J},
\end{align}
 
 In addition, a context normalization~\cite{pointcn} branch is used to predict inlier likelihood scores $\gamma$ for each seed correspondence, which will be used as weighting score for seed features in later unpooling stage.
 
\vspace{-1em}
\begin{align}
&\prescript{t}{}{\gamma}=\textrm{CN}(\mathbf{\prescript{t}{}{S}}_\mathbf{A}^5||\mathbf{\prescript{t}{}{S}}_\mathbf{B}^5),
\end{align}
where $\textrm{CN}$ is a lightweight stacked context normalization~\cite{pointcn} blocks. Detailed structures of $\textrm{CN}$ brach can be found in Appendix A.2. 

The outputs of seed filtering are the filtered features $\mathbf{\prescript{t}{}{S}}_\mathbf{A}^5$ , $\mathbf{\prescript{t}{}{S}}_\mathbf{B}^5$ and inlier scores  $\gamma \in [0,1]^k$ of seed matches.

\smallskip\noindent\textbf{Attentional unpooling.}
After the message exchange between seed matches and inlier score prediction, an inlier-score-weighted attentional aggregation is adopted to broadcast the pooled contexts to every keypoint in each graph, which we refer to as \textbf{attentional unpooling}.

Taking  $\mathbf{\prescript{t}{}{S}}_\mathbf{A}^5$ , $\mathbf{\prescript{t}{}{S}}_\mathbf{B}^5$  , inlier score $\gamma$ and $\mathbf{\prescript{t}{}{F_A}}$ , $\mathbf{\prescript{t}{}{F_B}}$ as input, attentional unpooling outputs the updated keypoint features $\mathbf{\prescript{t+1}{}{F_A}}$ , $\mathbf{\prescript{t+1}{}{F_B}}$.
\begin{align}
\mathbf{\prescript{t+1}{}{F_I}}=\textbf{Att}\mathbf{(\prescript{t}{}{F_I}},\mathbf{\prescript{t}{}{S}}_\mathbf{I}^5,\prescript{t}{}{\gamma}), \ \mathbf{I \in \{A,B\}}
\end{align}
Applying inlier score to the aggregation process suppresses information broadcast from false seed matches and results in cleaner feature update, which contributes to the robustness of our network w.r.t. seeding noise (Appendix D.2, Appendix Fig~10).

\smallskip\noindent\textbf{Assignment matrix formulation.}
After all processing units, the updated features are used to construct the assignment matrix.
Sinkhorn~\cite{sink} algorithm is applied on the correlation matrix of features with a dustbin channel to produce final assignment matrix $\mathbf{M}$.

Given the keypoint features $\mathbf{\prescript{N}{}{F_A}} \in \mathbb R^{n \times d}$, $\mathbf{\prescript{N}{}{F_B}} \in \mathbb R^{m \times d}$ after $N$ processing blocks, we compute the assignment matrix by
\begin{align}
&\mathbf{M}=\text{Sinkhorn}(\mathbf{\hat{C}})\\
&\hat{C}_{i,j}=\left\{\begin{aligned}
&C_{i,j},\ \text{for} \ i\leqslant n, j\leqslant m\\
&z,\ \text{otherwise}
\end{aligned}\right., \mathbf{\hat{C}} \in \mathbb R^{(n+1) \times (m+1)}
\end{align}
where $\mathbf{C}=\mathbf{\prescript{N}{}{F_A}\prescript{N}{}{F_B}^T}$, $z$ is a learnable parameter for dustbin. We derive final matches from the assignment matrix $\mathbf{M}$ with a confidence threshold to remove outliers. 

\subsection{Reseeding}

Although Seeded GNN based on initial seeding exhibits strong capability to identify underlying matches, a second seeding module using updated feature provides even cleaner and richer seeds to further improve the performance. Thus, we adopt a reseeding module. Different from initial seeding where NN matches and ratio scores of raw descriptors are used, reseeding module employs assignment matrix $\mathbf{M}$ of updated features to regenerate seeds. More specifically, matches with highest score in both rows and columns are selected as candidates, where top-k matches are selected as new seeds and are fed into a second Seeded GNN for refinement. More details can be found in Appendix A.2.

\subsection{Loss}

Seeding module only outputs seed indices that requires no gradient back-propagation, thus our network is fully differentiable and can be trained end-to-end with supervision from indices of ground truth matches $\mathbf{l_m}=\{(i,j)\} \in \alpha \times \beta$ and unmatchable points $\mathbf{l_{uA},l_{uB}}$, where a point is regarded as unmatchable if there are no matchable points in the other image. From assignment matrix $\mathbf{M}_r$ for reseeding, final assignment matrix $\mathbf{M}_f$, and inlier scores $\prescript{t}{}{\gamma}, t \in \{1,2,...,L\}$ for $L$ processing units, we formulate our loss as two parts,

\begin{align}
L=L_{assign}+\delta\Sigma_{t \in \{1,2,...,L\}} L_{weight}^t
\end{align}
where
\begin{align}
L_{assign}=-&\sum_{\mathbf{M}\in \{\mathbf{M}_r,\mathbf{M}_f\} }[\sum_{(i,j)\in \mathbf{l_m}}\log(M_{i,j})+ \nonumber\\
&\sum_{i\in \mathbf{l_{uA}}}\log(M_{i,m+1})+ \sum_{j\in \mathbf{l_{uB}}}\log(M_{n+1,j})]
\end{align}

$L_{weight}^t$ is cross entropy loss for inlier/outlier binary classification in $t$-th processing unit, a seed correspondence is labeled as inlier if its epipolar distance is less than a threshold. $\delta$ is a weight to balance the two loss terms.
\begin{table*}[ht]
\centering
\resizebox{0.95\textwidth}{!}{
\begin{tabular}{c|c|cc|cc|cc|cc}
	\Xhline{1pt}
	\multirow{2}{*}{\textit{Feature}}  & \multirow{2}{*}{\textit{Matcher}} & \multicolumn{2}{c|}{\textbf{CPC}}                                                  & \multicolumn{2}{c|}{\textbf{T\&T}}   & \multicolumn{2}{c|}{\textbf{TUM}}  & \multicolumn{2}{c}{\textbf{KITTI}}                                                 \\ \cline{3-10} 
	&                          & \textit{\%Recall} & \textit{\#Corrs(-m)} & \textit{\%Recall} &  \textit{\#Corrs(-m)} & \textit{\%Recall} & \textit{\#Corrs(-m)} & \textit{\%Recall} &  \textit{\#Corrs(-m)} \\ \Xhline{1pt}
	\multirow{4}{*}{\textbf{RootSIFT}~\cite{rootsift}}    & NN+RT                 &       52.9             &         92 (123)             &       82.1            &              208 (287)     & 61.9 & 365 (438)   & 90.6   &  847 (928)       \\
	& OANet~\cite{oanet}                    &        58.6           &                119 (167)          &      84.7          &         219 (306)    &          62.3         &             454 (396)       &       89.0               &        773 (854)           \\
	& SuperGlue~\cite{superglue}                &        61.1           &            218 (466)               &         \textbf{86.8}          &     382 (767) &       65.9          &      655 (1037)              &        91.0           &       \textbf{1261 (1746)}            \\
	& SGMNet                   &           \textbf{62.0}        &           \textbf{248 (524)}             &          85.9        &           \textbf{397 (789)}    &         \textbf{66.6}          &           \textbf{704 (1132)}              &        \textbf{ 91.2}         &            1097 (1506)            \\ \hline
	\multirow{4}{*}{\textbf{ContextDesc}~\cite{contextdesc}} & NN+RT                 &         62.4          &               169 (277)          &        85.5           &              222 (426)        &          58.7         &        456 (625)            &       90.6            &       1134 (1416)                  \\
	& OANet                    &         65.3          &      187 (288)           &         86.7          &         294 (425)     &       53.2            &         295 (327)             &        89.0            &           791 (907)         \\
	& SuperGlue                &      67.0          &         260 (579)            &        89.1          &      491 (695)      &          60.1        &          408(690)     &           \textbf{91.1}        &           \textbf{1401 (1897)}        \\
	& SGMNet                   &          \textbf{70.8}         &         \textbf{370 (616)}             &       \textbf{89.9}            &         \textbf{514 (705)}    &       \textbf{61.6}            &      \textbf{423 (705)}            &       90.3            &     1204 (1724)            \\ \hline
	\multirow{5}{*}{\textbf{SuperPoint}~\cite{sp}}  & MNN                 &       34.5            &       152 (421)               &          72.8         &             287 (717)     &          56.7         &             280 (420)            &         88.6          &       848 (1490)                \\
	& OANet                    &        62.9           &        186 (343)              &        91.2           &     280 (477)      &          61.4        &        332 (473)          &       82.2          &          482 (736)         \\

	& SuperGlue                &         68.8          &      287 (719)           &       92.9            &             414 (987)       &        59.1           &          512 (1038)              &          88.7      &      \textbf{957} (1777)      \\ 
	& SuperGlue$^*$               &        \textbf{76.7}          &      302(712)           &         \textbf{96.6}       &      431(985)              &        59.0           &         121(177)            &          \textbf{89.5}         &      354(526)     \\ 
	& SGMNet                   &         70.3      &         \textbf{327 (829)}             &            93.2      &          \textbf{450 (1098)}      &      \textbf{65.8}           &          \textbf{666 (1315)}              &         86.3         &       954(\textbf{1851})             \\ 
	\Xhline{1pt}
\end{tabular} 
}
\caption{Evaluation results on FM-Bench~\cite{fmbench}, where \textit{\%Recall} denotes mean recall of all pairs,  \textit{\#Corrs(-m)} denotes mean number of inlier correspondences after/before RANSAC. SuperGlue$^{*}$ indicates the results obtained from officially released model.}
\vspace{-1em}
\label{fm} 
\end{table*} 
\subsection{Implementation Details}
We train our network on GL3D dataset~\cite{gl3d}, which covers both indoor/outdoor scenes, to obtain general purpose model. We sample $1k$ keypoints and $128$ seeds during training. We use Adam optimizer with learning rate of $10^{-4}$ in optimization and inlier score weight $\delta$ in loss is set to 250. We use 6/3 processing blocks for initial/refine stage and the gradients flow between the two stages are blocked in early iterations(140k iterations). We use 4-head attention in both attentional pooling/unpooling operations. For all experiments, we use a confidence threshold of 0.2 to retain matches and seeding number of $\frac{128\#keypoint}{2000}$, where $\#keypoint$ is the number of keypoints. More details including training data generation and hyper-parameters can be found in Appendix~A.

\let\saveFloatBarrier\FloatBarrier
\let\FloatBarrier\relax

\section{Experiments}
\let\FloatBarrier\saveFloatBarrier
In the following sessions, we provide experiments results of our methods under a wide range of tasks, as well as further analysis of its computation and memory efficiency.

\subsection{Image Matching}
\smallskip\noindent\textbf{Datasets.} The performance of our method is first evaluated on image matching tasks and three benchmarks in two-view pose estimation, YFCC100M~\cite{yfcc}, FM-Bench~\cite{fmbench} and ScanNet~\cite{scannet} dataset, are used for demonstration. 

For YFCC100M~\cite{yfcc}, we follow the setting in OANet~\cite{oanet} and choose 4 sequences for testing. FM-Bench~\cite{fmbench} comprises four subsets in different scenarios: KITTI~\cite{kitti} for driving settings, TUM~\cite{tum} for indoor SLAM settings, Tanks and Temples~(T\&T)~\cite{tt} and CPC~\cite{cpc} for wide-baseline reconstruction tasks. ScanNet~\cite{scannet} is a widely used indoor reconstruction dataset. Following SuperGlue~\cite{superglue}, we use 1500 pairs in test set for evaluation. 

\smallskip\noindent\textbf{Evaluation protocols.} On YFCC100M and ScanNet dataset, pose estimation is performed on the correspondences after RANSAC post-processing. We report 1) AUC~\cite{superglue,oanet,pointcn} under different thresholds, computed from the angular differences between ground truth and estimated vectors for both rotation and translation; 2) Mean matching score (\textit{M.S.})~\cite{superglue,sp}, the ratio of correct matches and total keypoint number; 3) Mean precision (\textit{Prec.})~\cite{superglue,sp} of the generated matches. We detect up to $2k$ keypoints for all features on YFCC100M, up to $1k$ keypoints for superpoint on ScanNet and up to $2k$ keypoints for other features  .

On FM-Bench dataset, we estimate fundamental matrix for each evaluated pair with RANSAC post-processing, and use the normalized symmetric epipolar distance (SGD)~\cite{sgd,fmbench} as originally defined in FM-Bench paper, to measure the difference between the estimated fundamental matrix and the ground truth. An estimate is considered correct if its normalized SGD to ground truth is lower than a threshold ($0.05$ is used by default), and up to $4k$ keypoints are detected for each test pair. Following FM-Bench paper~\cite{fmbench} , we report: 1) recall (\textit{\%Recall}) on fundamental matrix estimation; 2) mean number of correct correspondences (\textit{\#Corrs(-m)}) after/before RANSAC.

\begin{table}[t]
	\centering
    \resizebox{0.48\textwidth}{!}{
	\begin{tabular}{c|c|ccc|c|c}
		\Xhline{1pt}
		\multirow{2}{*}{\textit{Feature}}  & \multirow{2}{*}{\textit{Matcher}} & \multicolumn{3}{c|}{\textbf{AUC}}                                                       & \multirow{2}{*}{\textbf{M.S.}} & \multirow{2}{*}{\textbf{Prec.}} \\ \cline{3-5}
		&                          & \textit{@5\textdegree}     & \textit{@10\textdegree}     & \textit{@20\textdegree}                   &                    \\ \Xhline{1pt}
		\multirow{4}{*}{\textbf{RootSIFT}}    & NN + RT~\cite{lowe2004distinctive}                 &    49.07      &     58.76      &    68.58            &         8.23          &      29.79            \\ & AdaLAM(4k)~\cite{adalam}                 &    57.78      &    68.01      &    77.38            &         7.92          &      83.15\\
		& OANet~\cite{oanet}                    &    58.00      &      67.80     &     77.46              &        5.84            &       81.80            \\ & SuperGlue$^{*}$~\cite{superglue} &   59.25   &  70.38  &    80.44   & - &-\\
		& SuperGlue~               &     \textbf{63.82}     &    \textbf{    73.33}  &    \textbf{82.26}                   &            16.59        &           81.08        \\  
		& SGMNet                   &     62.72     &      72.52     &    81.48               &     17.08             &          \textbf{86.08}          \\ \Xhline{1pt}
		\multirow{4}{*}{\textbf{ContextDesc}} & NN + RT                 &     57.90     &     68.47      &   78.35               &           9.39      &          59.72         \\ & AdaLAM(4k)~\cite{adalam}                 &    60.75      &    70.91     &     80.23         &         9.12         &      85.45 \\  
		& OANet                    &     62.28     &      72.56     &    81.80         &             9.33       &        \textbf{88.49}            \\  
		& SuperGlue                &      65.98    &     75.17      &     83.64            &        20.38            &      82.95              \\  
		& SGMNet                   &       \textbf{66.63}   &     \textbf{76.21}               &        \textbf{84.33}                    &        \textbf{20.57}            &       87.34             \\ \Xhline{1pt}
		\multirow{5}{*}{\textbf{SuperPoint}}  & MNN                 &   31.05       &    40.85      &    52.64               &          15.12          &           24.64         \\  & AdaLAM(2k)~\cite{adalam}                 &    40.20      &    49.03   &     59.11       &         10.17         &      72.57 \\
		& OANet                    &     48.80     &      59.06     &     70.02             &          12.48          &           71.95         \\  
		& SuperGlue$^{*}$                &      \textbf{67.10}    &  \textbf{76.18}        &     \textbf{84.37}                 &     21.58               &            \textbf{88.64}        \\  
		& SuperGlue                &     60.37     &     70.51      &     80.00          &            19.47        &           78.74       \\  
		& SGMNet                   &    61.22     &      71.02    &    80.45                 &           \textbf{22.36}         &        85.44           \\ \Xhline{1pt}
		\multirow{4}{*}{\textbf{SIFT}} 
		&PointCN~\cite{pointcn}&47.98&58.13&68.67&-&-\\
		&ACNe~\cite{sun2020acne} &-&-&78.00&-&- \\
		&LGLFM~\cite{darmon2020learning} &49.60&60.36&71.37&-&-\\ \Xhline{1pt}
		-&RANSAC-Flow~\cite{shen2020ransac}&64.88&73.31&81.56&-&-\\ \Xhline{1pt}
	\end{tabular}
	}
\caption{Results on YFCC100M~\cite{yfcc}, where AUC evaluates the pose accuracy, $M.S.$ denotes mean matching score and $Prec.$ denotes mean matching precision. SuperGlue$^{*}$ indicates the results obtained from original paper or officially released model.}
\vspace{-1em}
\label{yfcc}  
\end{table}
\begin{table}[t]
	\centering
    \resizebox{0.48\textwidth}{!}{
	\begin{tabular}{c|c|ccc|c|c}
		\Xhline{1pt}
		\multirow{2}{*}{\textit{Feature}}  & \multirow{2}{*}{\textit{Matcher}} & \multicolumn{3}{c|}{\textbf{AUC}}                                                       & \multirow{2}{*}{\textbf{M.S.}} & \multirow{2}{*}{\textbf{Prec.}} \\ \cline{3-5}
		&                          & \textit{@5\textdegree}     & \textit{@10\textdegree}     & \textit{@20\textdegree}                   &                    \\ \Xhline{1pt}
		\multirow{4}{*}{\textbf{RootSIFT}}    & NN + RT~\cite{lowe2004distinctive}                 &    9.08     &    19.75     &    32.66            &         2.28          &      28.83            \\ & AdaLAM~\cite{adalam}                 &       8.24   &     18.57    &         31.01       &            3.10    &    47.59  \\
		& OANet~\cite{oanet}                    &     10.71       &    23.10       &     37.42      &           3.20         &  36.93               \\  
		& SuperGlue~               &    \textbf{13.12}      &      \textbf{27.99}    &          43.92           &           \textbf{8.50}     &          42.53        \\  
		& SGMNet                   &    12.82     &      27.92    &        \textbf{44.55}        &        8.79            &          \textbf{45.55}        \\ \Xhline{1pt}
		\multirow{4}{*}{\textbf{ContextDesc}} & NN + RT                 &    11.07      &   23.52       &   37.66              &       5.29    &        28.71          \\ & AdaLAM               &    8.45     & 19.81       &          33.11     &      6.58         &    44.08   \\  
		& OANet                    &   11.95       &   24.49      &     40.56   &    5.12               &    40.43           \\  
		& SuperGlue                &     \textbf{15.70}    &   \textbf{31.67}       &       48.22        &       \textbf{10.75}             &          42.83        \\  
		& SGMNet                   &       15.46   &     31.55              &        \textbf{48.64}                    &        9.99            &       \textbf{48.14}            \\ \Xhline{1pt}
		\multirow{5}{*}{\textbf{SuperPoint}}  & MNN                 &   9.44       &    21.57      &    36.41              &          13.27          &           30.17         \\  & AdaLAM                &  6.72      &    15.82   &     27.37       &          13.19        &   44.22   \\
		& OANet                    &    10.04     &     25.09     &       38.01          &                10.56   &      44.61       \\  
		& SuperGlue                &     13.95     &     29.48      &    46.07       &       15.82           &       44.18        \\  & SuperGlue$^*$               &    \textbf{16.19}   &     \textbf{33.82}      &    \textbf{51.86}        &       \textbf{18.50}          &     47.32         \\
		& SGMNet                   &  15.40    &      32.06    &   48.32            &       16.97   &    \textbf{48.01}      \\    \Xhline{1pt}
	\end{tabular}
	}
\caption{Results on Scannet~\cite{scannet}. SuperGlue$^{*}$ indicates the results obtained from officially released model.}
\vspace{-1em}
\label{yfcc}  
\end{table}

\begin{figure*}[th]
	\centering
	\includegraphics[width=0.95\textwidth]{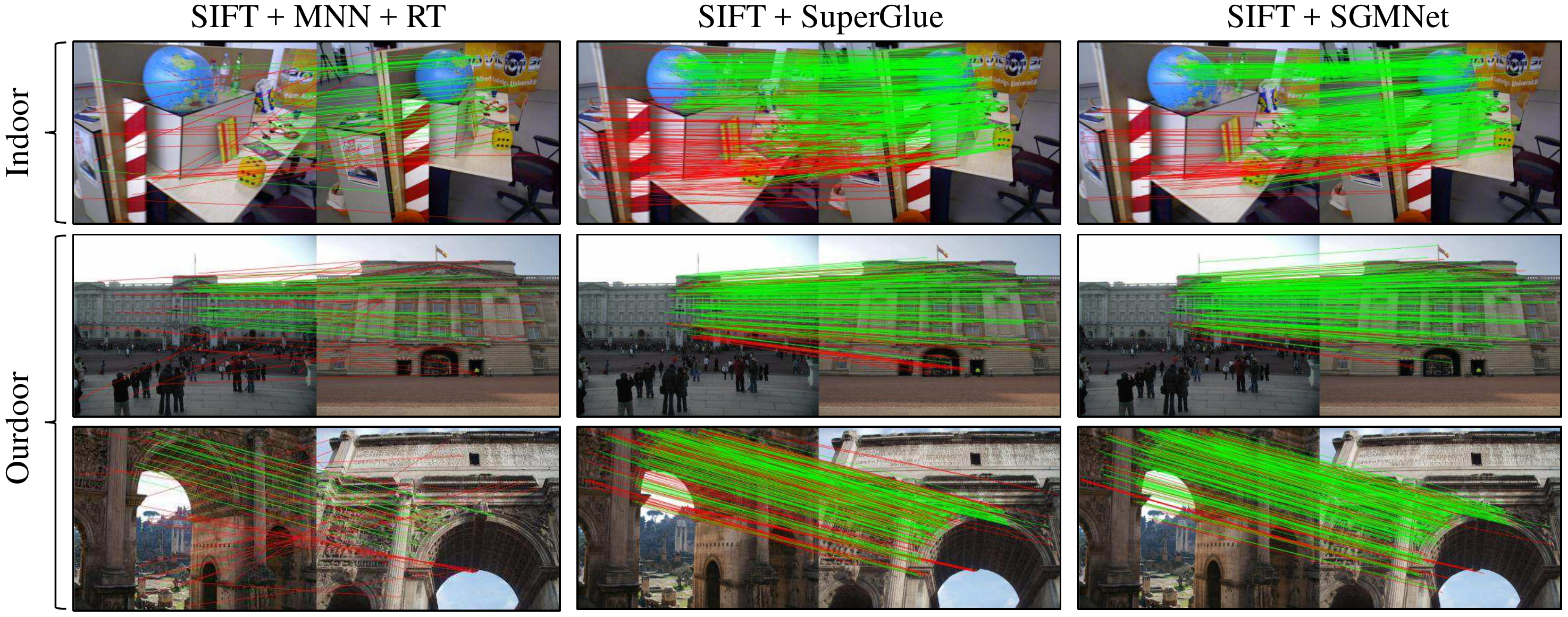}
	\caption{Correspondence visualizations. We showcase with SIFT features, and compare the results of traditional matching (\textit{MNN} + \textit{RT}), SuperGlue and our method (SGMNet). More visualizations available in Appendix. \vspace{-1em}}
	\label{fig:ret_vis}
\end{figure*}

\smallskip\noindent\textbf{Comparative methods.} We compare our method with heuristic pruning strategy, ratio test~\cite{lowe2004distinctive} or \textit{MNN}, and a various of learning-based matching methods~\cite{oanet,superglue,sun2020acne,darmon2020learning,pointcn,shen2020ransac}. These methods are applied on both handcrafted descriptors~\cite{lowe2004distinctive,rootsift} and learning-based local features~\cite{contextdesc,sp}.

For a fair comparison, OANet, SuperGlue and SGMNet are all re-trained using the same sequences of GL3D~\cite{gl3d}, where 1k keypoints are sampled per image. Noted that the official training code of SuperGlue is not available, and its public model (denoted as SuperGlue$^{*}$) is trained on MegaDepth~\cite{mega} and Oxford and Paris dataset~\cite{ox}. Instead, we retrain SuperGlue on GL3D~\cite{gl3d} with similar data selection criteria described in orignal paper. This re-implementation achieves even better results on YFCC100M (Table~\ref{yfcc}) for RootSIFT than those reported in the original paper. However, there still remains some performance gap when using SuperPoint~\cite{sp}, even though we have carefully tuned the training and enquired the authors about details. Nevertheless, we consider our re-implementation of SuperGlue~\cite{superglue} faithful and thus can be fairly compared. We report results of both official model and our re-implementation. 

\smallskip\noindent\textbf{Results.} For YFCC100M, ScanNet and two wide-baseline datasets of FM-Bench (CPC and T\&T), our method mostly shows competitive results compared with the state-of-the-arts. For two small-baseline datasets in FM-Bench (TUM and KITTI), the advantages of all learnable methods tend to degenerate due to the reduced matching difficulty. Our method matches most inlier correspondences on almost all dataset regarding \textit{M.S.} on YFCC100M/ScanNet and \textit{\#Corrs(-m)} on FM-Bench while maintaining a high matching precision, which contributes to the final pose accuracy. Though not specially trained in indoor scenarios, our method generalizes well on indoor setting.

\subsection{Visual Localization}
To evaluate how our method benefits real downstream applications, we integrate it in a visual localization pipeline and evaluate its performance.

\smallskip\noindent\textbf{Datasets.} We resort to Aachen Day-Night dataset~\cite{aachen} to evaluate the effectiveness of our method on visual localization task. Aachen Day-Night consists of $4328$ reference images and $922$ ($824$ daytime, $98$ nighttime) query images. All images are taken in urban scenes.

\smallskip\noindent\textbf{Evaluation protocols.} We use the official pipeline of Aachen Day-Night benchmark. Correspondences between reference images are first used to triangulate a 3D reconstruction. Correspondences between each query and its retrieved reference images are then generated to recover the relative pose. Consistent with the official benchmark, we report the pose estimation accuracy under different thresholds. We extract $8k$ keypoints for RootSIFT, ContextDesc and $4k$ keypoints for SuperPoint.

\smallskip\noindent\textbf{Results.}
Compared with SuperGlue, our method exhibits better results when using RootSIFT and competitive results when using SuperPoint or ContextDesc. Our method consistently outperforms OANet using all three descriptors. The overall performance proves the generalization ability of our method on real challenging applications. 

\begin{table}[t]
	\resizebox{0.48\textwidth}{!}{
	\begin{tabular}{c|c|ccc}
		\Xhline{1pt}
		\textit{Feature}                   & \textit{Matcher}   & \textbf{0.25m, 2\textdegree} & \textbf{0.5m, 5\textdegree} & \textbf{5m, 10\textdegree} \\ \Xhline{1pt}
		\multirow{4}{*}{\textbf{RootSIFT}~\cite{rootsift}}    & MNN  &     43.9                        &           56.1                 &          65.3                 \\
		& OANet~\cite{oanet}     &         69.4                    &        83.7                    &      94.9                     \\
		& SuperGlue~\cite{superglue} &            63.3                 &        80.6                    &       \textbf{98.0}                    \\
		& SGMNet    &             \textbf{70.4}               &          \textbf{85.7}               &        \textbf{98.0} \\ \hline
		\multirow{4}{*}{\textbf{ContextDesc}~\cite{contextdesc}} & MNN  &          65.3                   &             80.6               &           90.8             \\
		& OANet     &           74.5                  &          86.7                  &           \textbf{99.0}                \\
		& SuperGlue &                \textbf{77.6}             &       86.7                   &      \textbf{99.0}                     \\
		& SGMNet    &              75.5               &          \textbf{87.8}                  &       \textbf{99.0}                    \\ \hline
		\multirow{5}{*}{\textbf{SuperPoint}~\cite{sp}}  & MNN  &              71.4               &              78.6              &            87.8               \\
		& OANet     &                77.6             &        86.7                    &            98.0               \\
		& SuperGlue$^*$ &          \textbf{79.6}                   &     \textbf{90.8}                       &        \textbf{100.0}                   \\
		& SuperGlue &              76.5      &               88.8            &      99.0
		\\
		& SGMNet    &           77.6                  &        88.8                    &            99.0               \\ \Xhline{1pt}
	\end{tabular}
	}
\caption{Evaluation results on Aachen Day-Night dataset. We report pose accuracy under different thresholds for challenging night spilt. We include the results of official released model of SuperGlue with SuperPoint (denoted as SuperGlue$^*$).}
\vspace{-1em}
\label{aa}
\end{table} 

\begin{figure}[t]
	\centering{\includegraphics[width=0.5\textwidth]{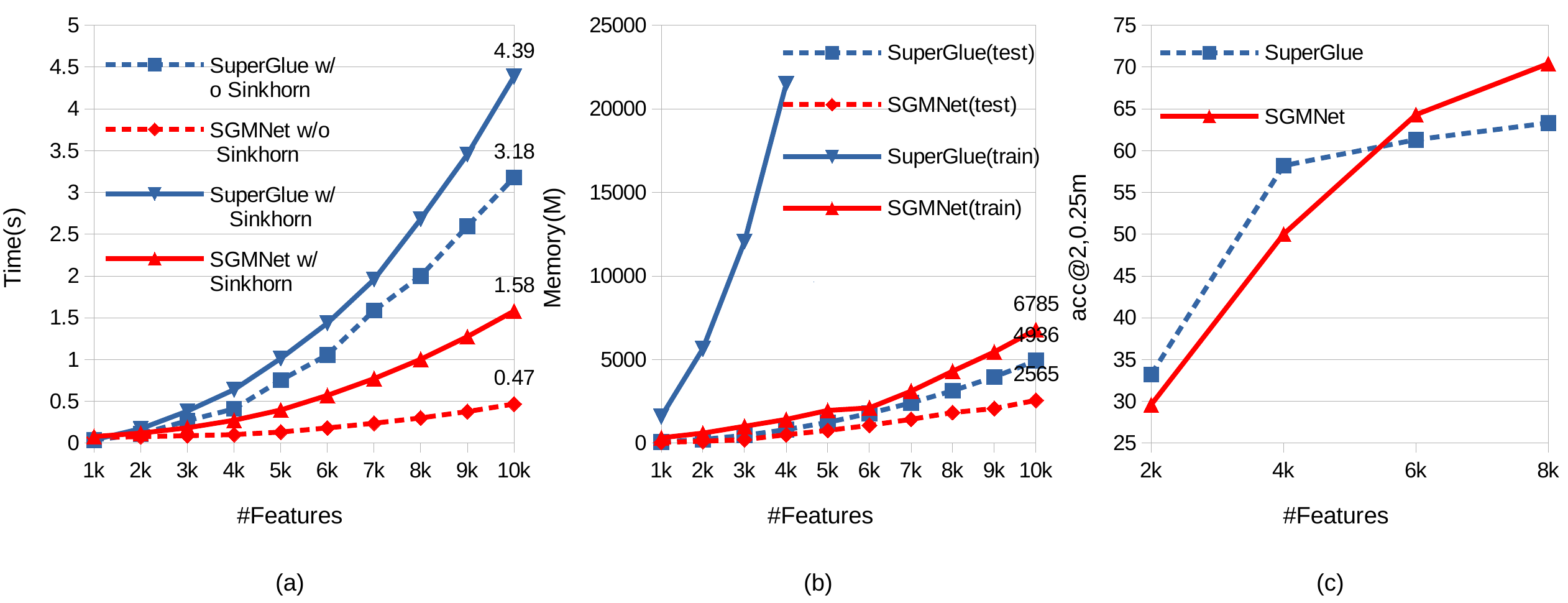}
	\caption{\label{fig:time} The computation (a) and memory (b) efficiency compared the proposed method with SuperGlue. We report memory occupation averaged by batch size for training. The effect of keypoint number on Aachen Day-Night dataset is illustrated in (c).}
	\vspace{-1.5em}
	\label{time}}
\end{figure}

\subsection{Scalability}

In above experiments, the proposed method has shown competitive results against the state-of-the-arts. In this session, we demonstrate the major advantage of our method in time/memory efficiency compared with SuperGlue, a closely related method based on GNN.

\smallskip\noindent\textbf{Time/memory consumption.} As shown in Fig.~\ref{time}(a), the time cost of our method is remarkably lower than SuperGlue. Specifically, we report both run time with and without Sinkhorn iterations on a GTX 1080 GPU, in order to more precisely demonstrate the substantial improvements on GNN design itself. It is noteworthy that on $10k$ keypoints and without Sinkhorn iterations, the proposed method reduces the runtime by one magnitude.  Besides, due to the reduction of redundancy, SGMNet also delivers better convergence during training (see Appendix~C). 

As shown in Fig.~\ref{time}(b), during test phase, our method consumes half of memory than SuperGlue when keypoint number is larger than $2k$, where the major memory peak of our method is seeding phase and sinkhorn iterations. The advantage becomes even more significant during training. With a batch size of 16 and keypoint number of $1k$, SuperGlue occupies up to 23GB GPU memory for training while SGMNet only consumes less than 9GB memory.

\smallskip\noindent\textbf{Performance gain when using more keypoints.} 
Within a reasonable range, larger keypoint number generally improves performance of downstream tasks, 
thus a manageable matching cost is of practical significance to extend the applicability. As a showcase, we vary the keypoint number of RootSIFT when evaluating on Aachen Day-Night Dataset. As can be seen from Fig.~\ref{time}(c), the accuracy of SGMNet and SuperGlue increases as more keypoints are used. Considering the efficiency advantage of our method, SGMNet delivers better trade-off when increasing the keypoint number. We also provide an experiment of SfM, a typical keypoint-consuming application in Appendix D.3.

\begin{table}[t]
\centering
\begin{tabular}{ll|ccc}
\Xhline{1pt}
\multicolumn{2}{c|}{\textit{Matcher}}            & \textit{AUC@20\textdegree} & \textit{M.S.} & \textit{Prec.} \\ \Xhline{1pt}
\multicolumn{2}{l|}{NN + RT}      &   68.58    &    10.05    &     56.38      \\
\multicolumn{2}{l|}{SGMNet w Rand. Seed} &    71.25  & 12.94       &     55.57    \\
\multicolumn{2}{l|}{SGMNet w/o W.U.}    &    78.64   &   17.07    &     81.26      \\
\multicolumn{2}{l|}{SGMNet w/o A.P.}    &    79.35  &    17.11    &     82.15      \\
\multicolumn{2}{l|}{SGMNet w/o Reseeding}        &    80.41    &   \textbf{17.12}    &      84.47      \\ \hline
\multicolumn{2}{l|}{SGMNet full}        &   \textbf{81.48}    &   17.08    &      \textbf{86.08}      \\\Xhline{1pt}
\end{tabular}
\caption{Results of ablation study. w/o A.P. stands for w/o attentional pooling, where seed features are directly sent to seed filtering process without attending to original keypoints. w/o W.U. stands for w/o weighted unpooling, where vanilla attention are preformed in unpooling process. Rand. Seed means selecting seed correspondences randomly instead of picking the top-k scores. w/o Reseeding means only use initial seeding.}
\vspace{-1.5em}
\label{ablation}
\end{table}

\section{Discussions}
\subsection{Ablation Study}
To evaluate the effectiveness of different components of our method, we conduct ablation study on YFCC100M dataset using RootSIFT. As shown in Table~\ref{ablation}, all different component in our network contributes to the final performance notably. In particular, seeding reliable matches plays an important role, which further proves that seed matches is able to guide message across images for robust matching.

\begin{table}[]
\resizebox{0.48\textwidth}{!}{
\begin{tabular}{c|c|ccc|c}
\Xhline{1pt}
\multirow{2}{*}{\textit{Type}} & \multirow{2}{*}{\textit{Matcher}} & \multicolumn{3}{c|}{\textit{AUC}}                           & \multirow{2}{*}{\textit{Time(ms)}} \\ \cline{3-5}
                         & & \textit{@5\textdegree}                    & \textit{@10\textdegree}                   & \textit{@20\textdegree} &                              \\ \Xhline{1pt}

\multirow{4}{*}{\textbf{GNN}} & {SGMNet-10}              & \multicolumn{1}{c|}{66.62} & \multicolumn{1}{c|}{76.33} &  \textbf{84.71}  &              114.81               \\
& {SGMNet}              & \multicolumn{1}{c|}{\textbf{67.42}} & \multicolumn{1}{c|}{\textbf{76.63}} &  84.66  &              284.66                \\
& {SuperGlue-10}              & \multicolumn{1}{c|}{65.85} & \multicolumn{1}{c|}{75.35} &  84.05  &              458.52              \\
& {SuperGlue}                & \multicolumn{1}{c|}{66.65} & \multicolumn{1}{c|}{75.41} &  84.12  &   604.11                         \\ \hline
\multirow{2}{*}{\textbf{Filter}} & OANet                    & \multicolumn{1}{c|}{63.75} & \multicolumn{1}{c|}{73.60} &  82.43  &                21.30              \\ 
& ACNe                    & \multicolumn{1}{c|}{63.37} & \multicolumn{1}{c|}{73.67} &   82.74 &                   18.26           \\ \hline
\multirow{2}{*}{\textbf{Handcrafted}} & AdaLAM                  & \multicolumn{1}{c|}{57.78} & \multicolumn{1}{c|}{68.01} &  77.38  &  4.59                            \\ 
& GMS                      & \multicolumn{1}{c|}{26.15} & \multicolumn{1}{c|}{33.53} &   42.13 &        5.44                      \\ \Xhline{1pt}
\end{tabular}
}
\caption{Results on YFCC100M using 4k SIFT features. -10 means setting sinkhorn iterations as 10 instead of 100}

\label{efficiency}  
\vspace{-1.5em}
\end{table}

\subsection{Comparison with Filter-based Methods} For a well-around comparison, we provide in Table~\ref{efficiency} more experiment results with filter-based methods (outlier rejection). When using 4k keypoints, SGMNet achieves best performance among all comparative methods while runs 4 times faster than SuperGlue when setting sinkhorn iterations as 10. Despite the fast inference speed of SOTA filter-based methods, a considerable performance gap still remains compared with GNN based methods.

\subsection{Designs of SGMNet}
We experiment on other designs for SGMNet, including other seeding strategy and pooling operations in GNN/Transformer architecture~\cite{transformer,transpool1,diffpool}, e.g. diffPool~\cite{oanet,diffpool} and set transformer~\cite{settrans} . We find 1) learnable seeding acheives limitied improvement over our simple heuristic seeding strategy 2) other general pooling operations, which are often verified on self-attention in GNN/Transformer architecture, are not effective alternatives of our seed-based pooling. Details of our experiments can be found in Appendix~B. Hyper-parameter study on seeding number can be found in Appendix~D.   
\vspace{-0.25em}
\section{Conclusion}
In this paper, we propose SGMNet, a novel graph neural network for efficient image matching. The new operations we have developed enable message passing with a compact attention pattern. Experiments on different tasks and datasets prove that our method improves the accuracy of feature matching and downstream tasks to competitive or higher level against the state-of-the-arts with a modest computation/memory cost.

\vspace{+0.5em}
\smallskip\noindent\textbf{Acknowledgment.} This work is supported by Hong Kong RGC GRF 16206819, 16203518, T22-603/15N and Guangzhou Okay Information Technology with the project GZETDZ18EG05.

\section*{Supplementary Appendix}

\section*{A Implementation Details}
We provide in this part details about our implementations as well as experiment settings.
\subsection*{A.1 Training Details}
\smallskip\noindent\textbf{Training Data.} We use GL3D dataset to generate training data. GL3D dataset is originally based on 3D reconstruction of 543 different scenes, including landmarks and small objects, while in its latest version additional 713 sequences of internet tourism photos are added.

We sample 1000 pairs for each sequence and filter out pairs that are either too hard or too easy for training. More specifically, we use the common track ratio provided by original dataset and rotation angles between cameras to determine pair difficulty, pairs with common track ratio in range $[0.1,0.5]$ and rotation angle in range $[6\textit{\textdegree},60\textit{\textdegree}]$ are kept. 

We reproject keypoints between images with depth maps and use reprojection distances to determine ground truth matches and unmatchable points. More specifically, a keypoint is labeled as unmatchable if its reprojection distances with all keypoints in the other image are larger than 10 pixels, while a pair of  keypoints that are mutual nearest after reprojection and with a reprojection distance lower than 3 pixels are considered ground truth matches. We further filter out pairs with ground matches fewer than 50.

Our data generation protocol yields around 400k training pairs in total.

\begin{figure*}[t]
	\centering
	\includegraphics[width=\textwidth]{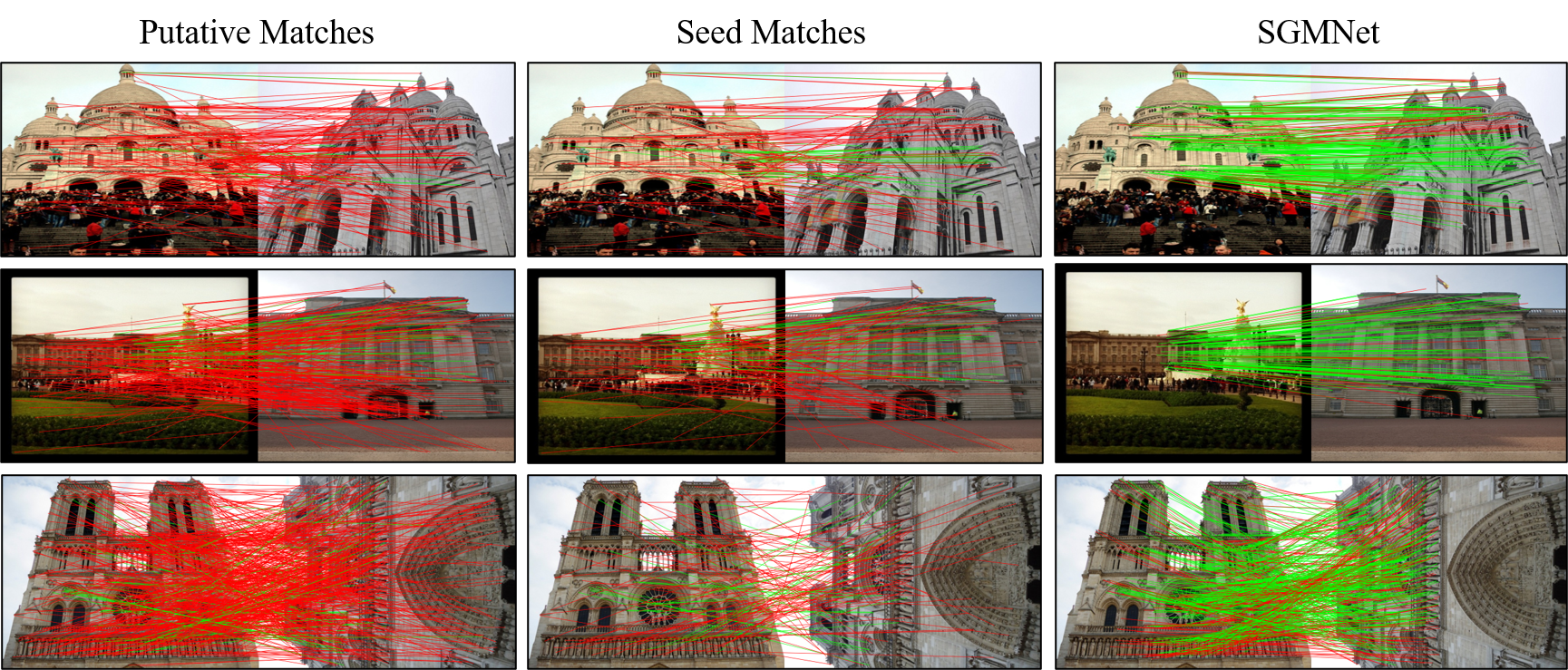}
	\caption{Visualizations of raw putative matches(left), seed correspondences(middle) and matches obtained by SGMNet(right). Note that even with heavily noisy seeds, SGMNet is capable of discovering underlying patterns, which are leveraged to guided message pass across keypoints for robust and accurate matching.  
	}
	\label{fig:seed}
\end{figure*}

\smallskip\noindent\textbf{Training Parameters.} We use Adam optimizer for training optimization with learning rate of $10^{-4}$. We apply learning rate decay after 300k iterations with decay rate of 0.999996 until 900k iterations. We block the gradient flow between initial/refinement stages for the first 140k iterations. 

\subsection*{A.2 Network Details}
As is done in SuperGlue~\cite{superglue}, we apply multi-head attention mechanism for all weighted attention operations in our network with head number of 4. 

For initial Seeding Module, we apply additional mutual nearest neighbour check and ratio test to seed correspondences. NMS is also employed in seeding. The radius for NMS is $\frac{\theta}{|\alpha|(|\alpha|-1)}\underset{(i,j)\in \alpha\times\alpha,i\neq j}{\Sigma}d_{ij}$, where $\alpha$ is index set for all keypoints, $d_{ij}$ is distance between keypoint $i,j$, $|.|$ denotes set size, $\theta$ is a hyper parameter, which we set as $10^{-2}$ for all experiment.  

For Reseeding Module, we apply sinkhorn algorithm with 10 iterations on correlation matrix to obtain assignment matrix. Correspondences with top-k scores on both dimensions are sampled as seeds. NMS is also applied in reseeding module.

For the inlier likelihood predictor $CN$ in Seeded GNN, we use the $PointCN$ structure illustrated in Fig~\ref{fig:cn} (down).

We set iterations number of Sinkhorn algorithm to 100.

\subsection*{A.3 Experiment Settings}
We use OpenCV implementation for SIFT and official implementation of SuperPoint and ContextDesc. For ContextDesc, we use the latest public model (denoted as \textit{ContextDesc++upright} in the official GitHub repository).

\smallskip\noindent\textbf{YFCC100M.} We extract SIFT and ContextDesc with images in original resolution and resize images so that the longest dimension is 1600 to extract SuperPoint. We use OpenCV \textit{findEssentialMat} and recoverPose functions to recover relative poses with the embedded RANSAC, of which
the threhold is set as 1 pixel under resolution of resized images. For matching score and precision, we use epipolar
distance of $5\times10^{-3}$ to determine inlier matches.

\smallskip\noindent\textbf{ScanNet.} We resize images to [640,480] resolution to extract keypoints and use same protocol to recover relative poses, mathing scores and precision as in YFCC100M.

\smallskip\noindent\textbf{FM-Bench.} The original evaluation pipeline of FM-Bench is based on Matlab and we reimplement it with Python. The parameter for evaluation is consistent with the original implementation. We use OpenCV $findFundamental$ function for fundamental matrix estimation and set the threshold of embedded RANSAC to 1 pixel. Compared with original implementation, our evaluation pipeline tends to give out higher accuracy, especially for wide baseline datasets. We believe it is due to better performance of OpenCV $findFundamental$ function and is beneficial for a more precise evaluation.

\smallskip\noindent\textbf{Aachen Day-Night.} We use the official pipeline and default parameters for evaluation. We extract upright feature for both RootSIFT and ContextDesc.

\section*{B Designs of SGMNet}
Despite the effectiveness of SGMNet, we provide our experiment results and analysis on some other potential designs in this part.
\begin{table}[t]
\resizebox{0.48\textwidth}{!}{
\centering
\begin{tabular}{c|ccc|c|c|c}
\hline\hline
                         & \multicolumn{3}{c|}{\textbf{AUC}} & \multicolumn{1}{c|}{\multirow{2}{*}{\textbf{M.S.}}} & \multirow{2}{*}{\textbf{Prec.}}&\multirow{2}{*}{\textbf{Prec.(S)}} \\ \cline{1-4}
\textbf{Designs} & \textit{@5\textdegree} &\textit{@10\textdegree} & \multicolumn{1}{c|}{\textit{@20\textdegree}} & \multicolumn{1}{c|}{}                               &                                 \\ \hline
Learned Seeding        & 62.80  &  72.55  & 81.09 &  17.25             &         85.15 & 51.22                        \\ \hline
DiffPool                 & 50.85  & 60.50   & 69.68   &         10.18     &     61.52  &-                          \\
ISA                & 52.11  & 61.44   & 70.03 &            10.65     &   63.24    &-                          \\ \hline
SGMNet                   &62.72   & 72.52   & 81.48&    17.08           &  86.08        &     39.24                  \\
NN+RT                    &  49.07 & 58.76   & 68.58&        10.05                  &         56.38&-                        \\ \hline\hline
\end{tabular}}
\caption{Evaluation Results on YFCC100M using RootSIFT with different designs. $\textbf{Prec.(S)}$ denotes precision of seed correspondences.}
\label{yfcc_design}  
\end{table}

\begin{figure}
	\centering
	\includegraphics[width=0.45\textwidth]{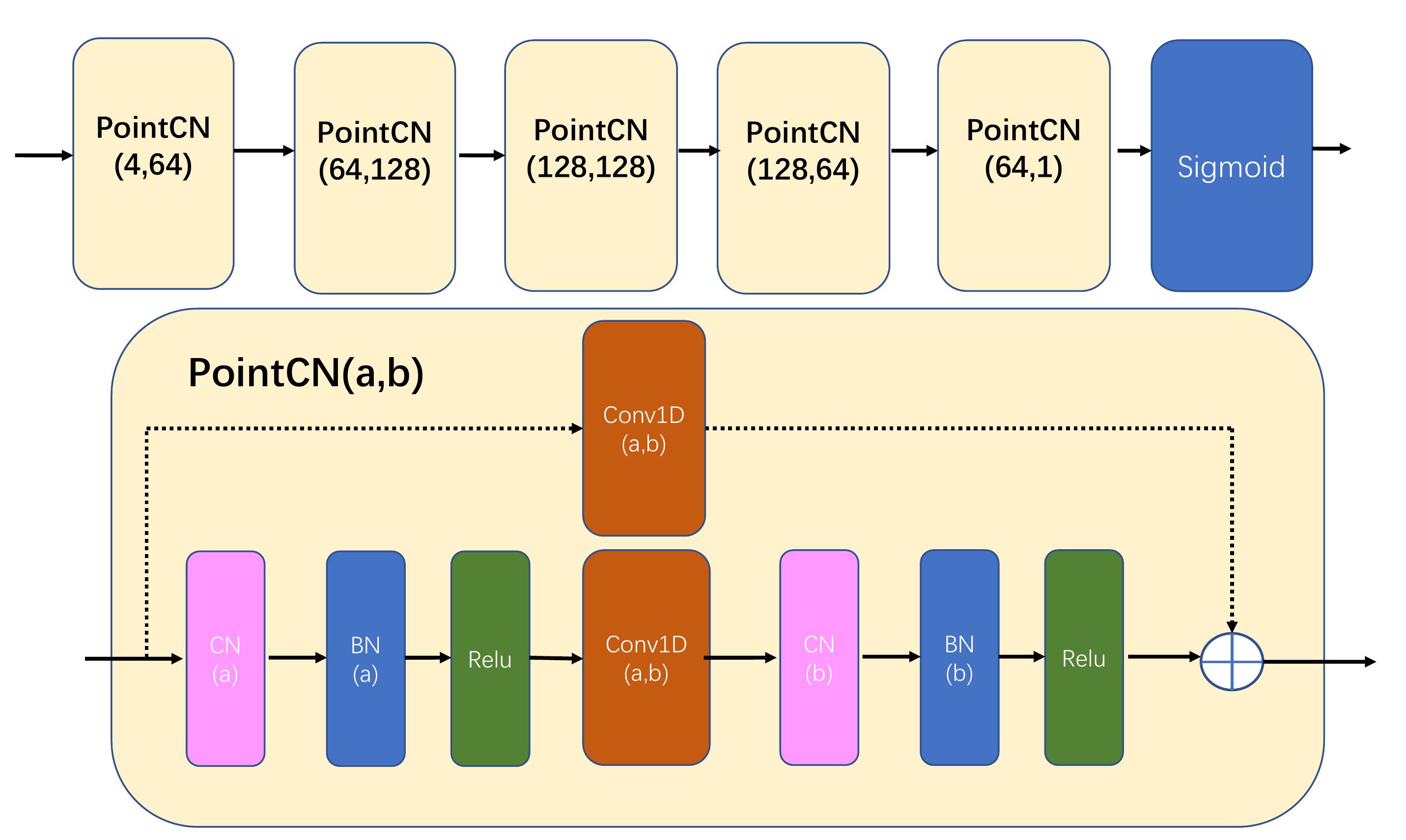}
	\caption{PointCN structure(down) we use to construct learned seeding module(top). $CN(a)$/$BN(a)$ denotes context normalization~\cite{pointcn}/batch normalization~\cite{bn} with input of $a$ channels.}
	\label{fig:cn}
\end{figure}

\smallskip\noindent\textbf{Learned Seeding.} After obtaining initial nearest neighbour correspondences, we employ a light-weight permutation invariant Network (architecture illustrated in Fig~\ref{fig:cn}) to determine each correspondence's inlier likelihood score, as is done in previous works~\cite{oanet,pointcn}. We thus sample correspondences with top-k inlier-score instead of ratio scores as seeding correspondences.

\smallskip\noindent\textbf{Results.} We report results in Table~\ref{yfcc_design}. Although applying light-weight pointCN block for inlier seeds prediction slightly increases seeding precision, it's not enough to bring meaningful impacts on the level of pose estimation. 

In general, we are open for the possibility to increase matching quality by introducing more complicated seeding strategies. However, targeting at efficient matching, our seeding method achieves good balance between performance and cost.

The critical component in our method for efficiency message passing is essentially pooling of original keypoints. In this part, we apply two well-studied pooling designs either in GNN or transformer architecture, namely DiffPool~\cite{diffpool} and Induced Set Attention~\cite{settrans}, to image matching task and evaluate their performance.

\smallskip\noindent\textbf{DiffPool.} As a pooling operation in GNN, DiffPool predicts assignment matrix based on each node's embedding in graph, which is designed to build hierarchical and sparse graph representation~\cite{diffpool}. In OANet~\cite{oanet}, DiffUnpool, the counter part of DiffPool, is proposed to recover node clusters to original size. As an experiment, we apply DiffPool/Unpool to keypoint graph in each image as a substitution for our proposed attentional pooling. Cross/self attention~\cite{superglue} will be performed on the pooled clusters for message exchange.

\smallskip\noindent\textbf{Induced Set Attention.} Induced set attention(ISA) is first proposed in Set Transformer~\cite{settrans}. Different from using seed features as attention bottleneck, ISA adopts a set of learned fixed features(induced point) as attention pass between set elements and is only verified on self-attention for sparse input. We subsitute our seed-based attention with ISA. More specifically, we let the network learned induced points for both sides and let induced points attend to original keypoints(both cross/self) to perform message pass.

\smallskip\noindent\textbf{Results.} We report evaluation result on YFCC100M. For all pooling method, we set pooling number to 128 and extract up to $2k$ keypoints. As illustrated in Table~\ref{yfcc_design}, both DiffPool and ISA shows only marginal improvements over baselines, which indicates that applying pooling methods designed for generic GNN/efficient transformer is not necessarily effective for image matching tasks, and further prove that our seed based attentional pooling/unpooling operation is critical for the success of our method.

\section*{C Fast Convergence}
\begin{figure}[h]
	\centering
	\includegraphics[width=0.3\textwidth]{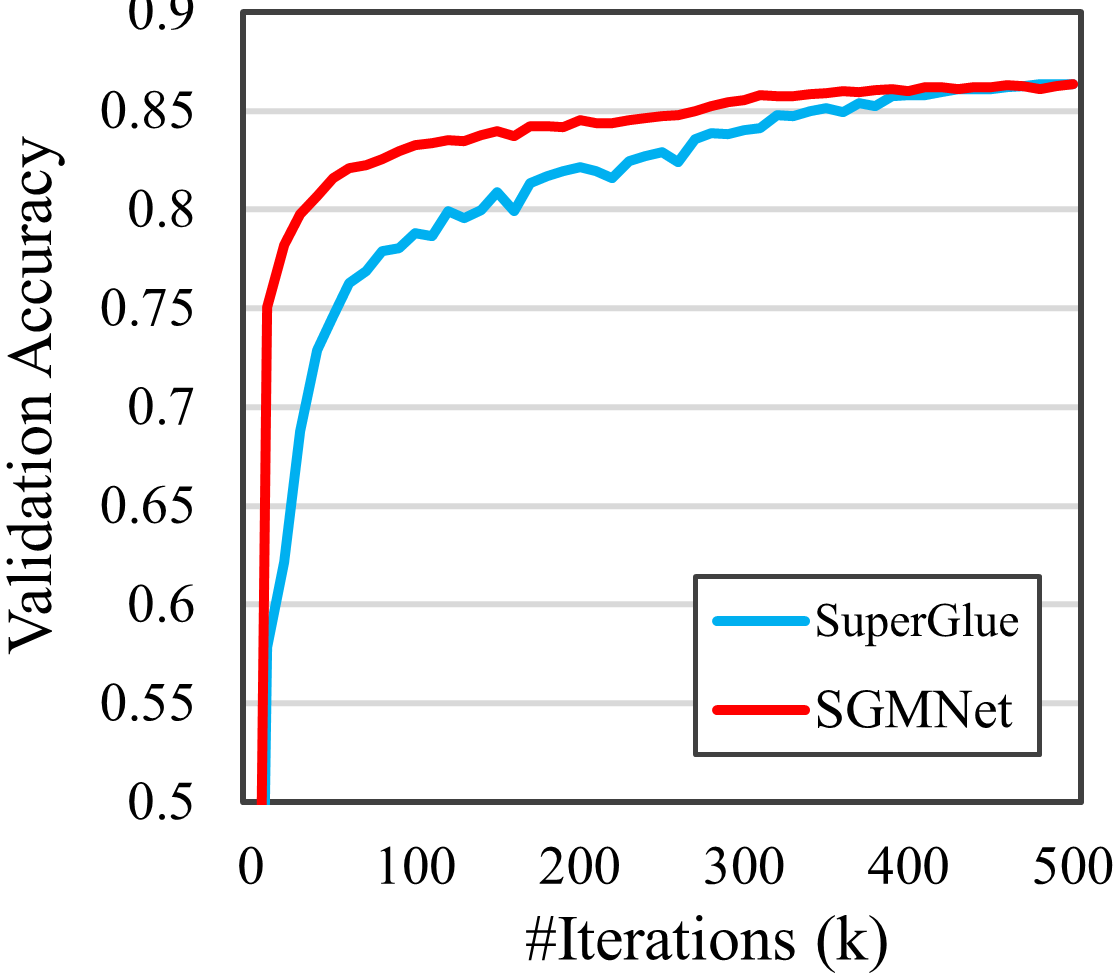}
	\caption{Convergence curve.
	}
	\label{fig:convergence}
\end{figure}

The compactness of SGMNet not only contributes to the cut-down on computation/memory complexity  but also leads to a faster convergence for training. In Fig.~\ref{fig:convergence} we plot the training curve for both SGMNet and SuperGlue. As illustrated, SGMNet takes fewer iterations to reach convergence.

\section*{D Additional Experiment Results}

\begin{figure}[t]
	\centering{\includegraphics[width=0.42\textwidth]{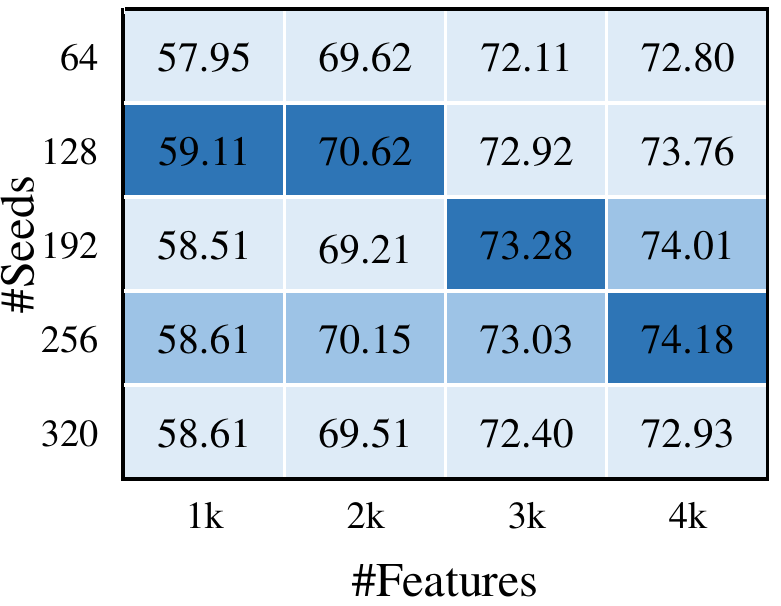}}
	\caption{The effect of seed number when varing the keypoint number. Numbers in grids are Exact AUC\textit{@20\textdegree} using RootSIFT. }
	\label{seed_num}
\end{figure}

\begin{figure}[th]
	\centering
	\includegraphics[width=0.4\textwidth,height=0.3\textheight]{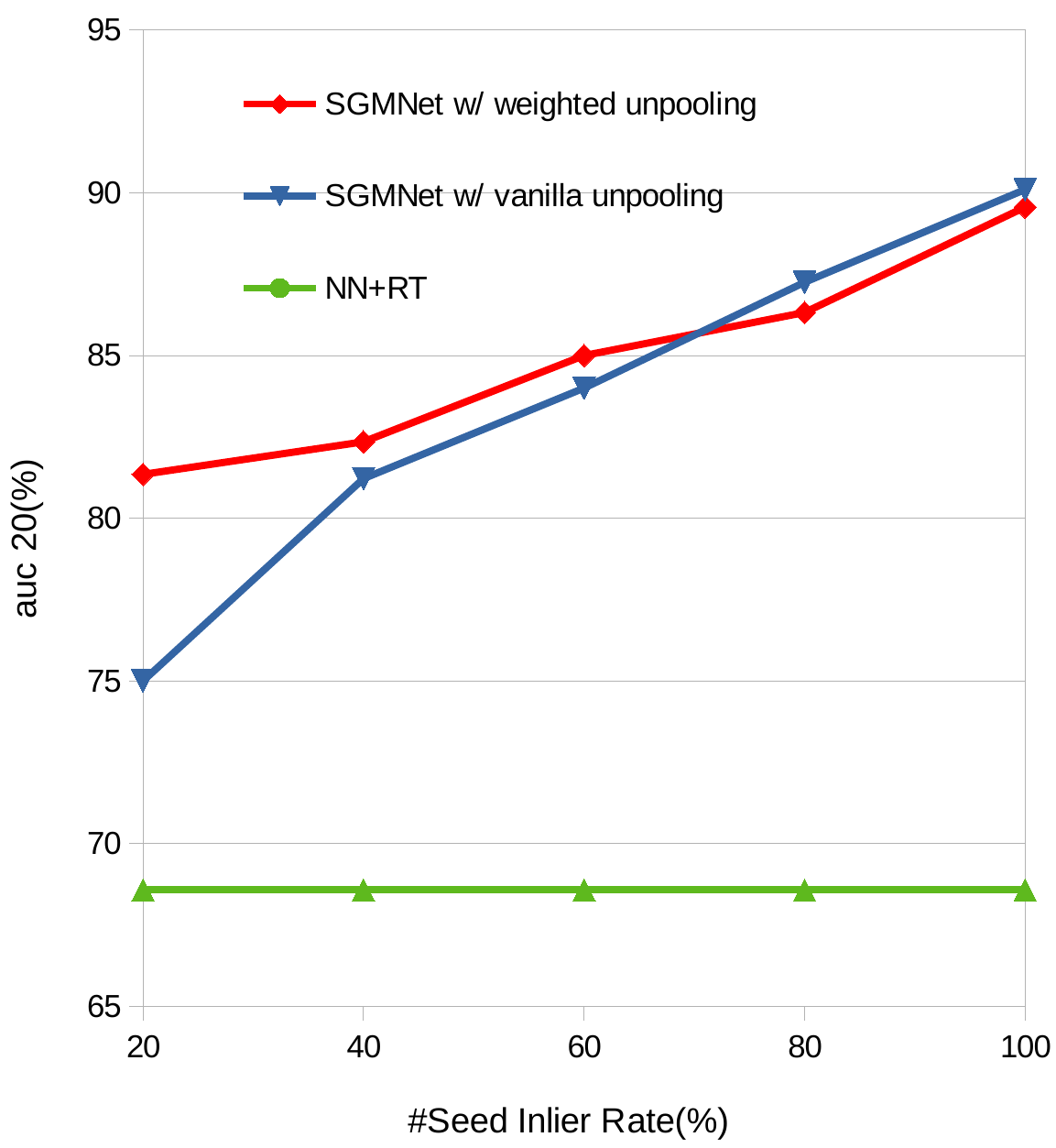}
	\caption{Relationship between pose estimation accuracy and seed precision. Note SGMNet maintain a high matching quality even with seed precision of only 20\%. 
	}
	\label{seed_precision}
\end{figure}

\begin{figure*}[t]
	\centering
	\includegraphics[width=1\textwidth,height=0.4\textheight]{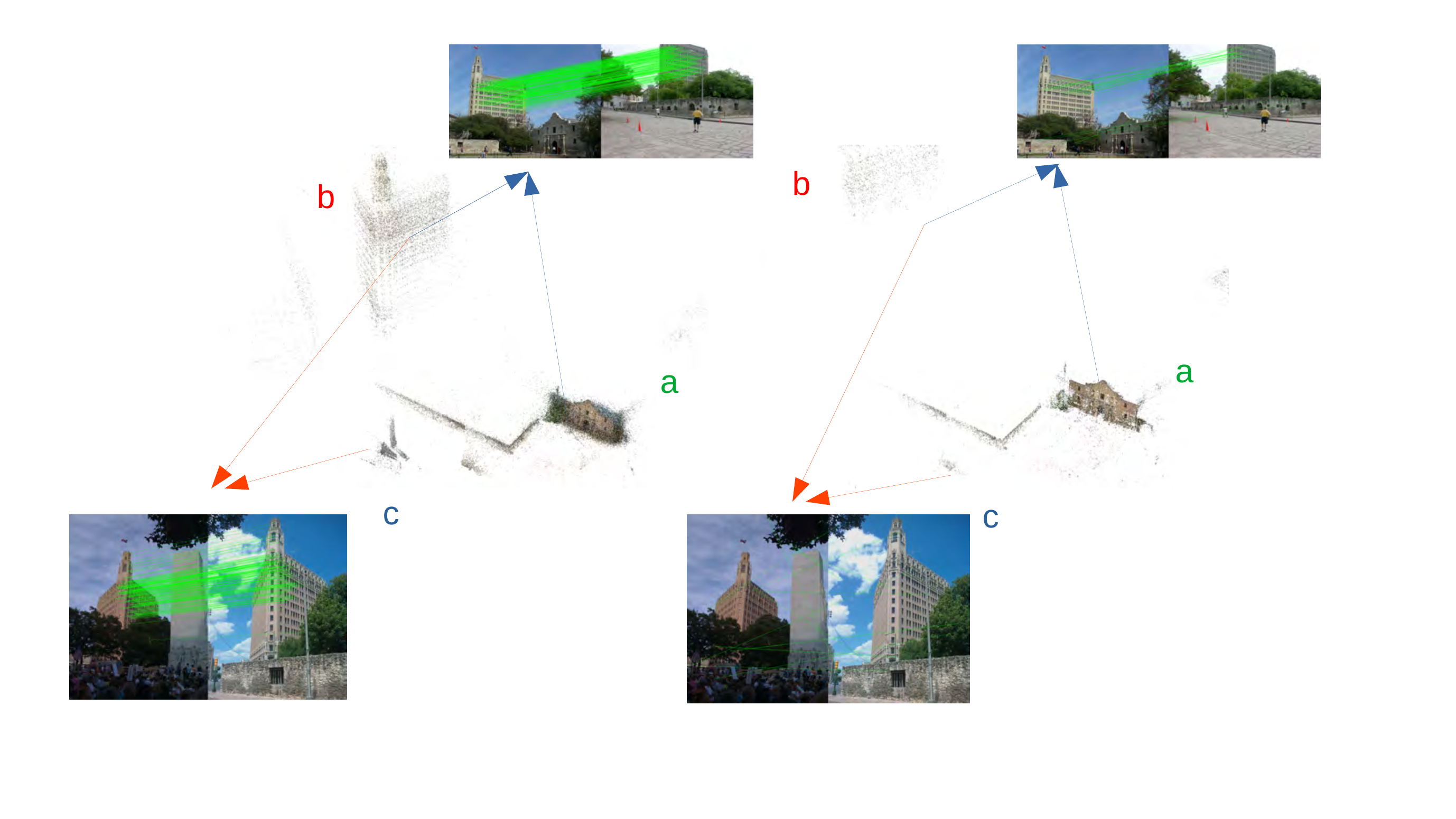}
	\caption{Reconstruction results of vanilla nearest neighbour matching(left) and SGMNet(right). The completeness of reconstruction is determined by matching quality between some critical frames. In this case, NN matching fails to generate descent correspondences between tall building(b)/statue(c) and tall building(b)/remains(a), which results in incomplete reconstruction, while SGMNet registered theses critical frames successfully.}
	\label{fig:more_vis}
\end{figure*}

\subsection*{D.1 Impact of Seeding Number}
SGMNet requires seeding a set of seed correspondences, the number of which not only influences our method's efficiency but also accuracy. Therefore, it is important to investigate the impact of seeding number. We carefully conduct grid search on YFCC100M with different keypoint and seeding number.

As is illustrated in Fig~\ref{seed_num}, an approximate proportional relationship between keypoint/seed numbers yields best performance, as too many seeds may deliver less reliable guidance while seeding too few correspondences results in severe information lost. 

\begin{table*}[t]
\resizebox{1\textwidth}{!}{
\centering
\begin{tabular}{c|c|c|c|c|c|c}
\hline\hline
\textbf{Methods}&\textbf{\#Registered Images}&\textbf{\#Sparse Points}&\textbf{Mean Rro. Error} &\textbf{Mean Track Len.} &\textbf{Matching Time}&\textbf{Total Time}\\
\hline
NN+Ratio+Mutual Check &799 &132265 & \textbf{0.58px}  &\textbf{11.27} & 1h 13min& 2h 22min  \\ 
SuperGlue &916 & 223950& 0.95px&10.93 &42h 34min &44h 06min\\
SGMNet & \textbf{943} & \textbf{276240}& 1.10px& 10.73 & 5h 37min & 7h 56min\\\hline\hline
\end{tabular}}
\caption{SfM results for Alamo scene in 1DSFM dataset.}
\label{sfm}  
\end{table*}

\subsection*{D.2 Robustness to Seed Noise}
To evaluate the robustness of our method w.r.t. potential false seeds, we conduct experiment on YFCC100M. More specifically, for each pair we select a set of inlier matches, which is determined using ground truth, and pad them with random sampled noise to construct seed correspondences with different precision. We feed the pre-selected seed correspondences to Seeded GNN instead of applying Seeding Module.  

As is shown in Fig~\ref{seed_precision},  SGMNet maintains high matching quality even with heavily noisy seed correspondences. It's noteworthy that for SGMNet without weighted unpooling, the pose estimation accuracy degenerate more rapidly as seed precision decreases, which indicates lower robustness to seed noise. More visualizations related to noisy seed and matching results can be seen in Fig~\ref{fig:seed}.

\subsection*{D.3 SfM Experiment}

Typical Structure from Motion(SfM) pipeline usually involves extracting keypoints in large number(e.g. 8k) and matching among hundreds/thousands of images to obtain ultra accurate poses and more complete reconstructions. In this section, we embed different matching methods into COLMAP SfM pipeline for comparison. We reconstruct challenging Alamo scene from 1DSFM~\cite{cpc}, which involves 2915 images taken under very different illumination conditions. 8k RootSIFT features are extracted for each image and we set sinkhorn iterations to 10 for both SGMNet and SuperGlue. We use a GTX 1080 GPU to preform matching sequentially. Apart from common statistics for reconstruction, we also report time consumption for matching and the whole SfM pipeline.

As shown in Tab~\ref{sfm}, both SuperGlue and SGMNet produces much more complete reconstruction compared with vanilla NN matching and heuristic pruning. However, SuperGlue largely lengthen the time for whole SfM pipeline, while our method retains the matching time to a feasible level.

\section*{E More Visualizations}
See Fig.~\ref{fig:more_vis} and Fig.~\ref{fig:seed}.
\clearpage

\begin{figure*}[t]
	\centering
	\includegraphics[width=\textwidth]{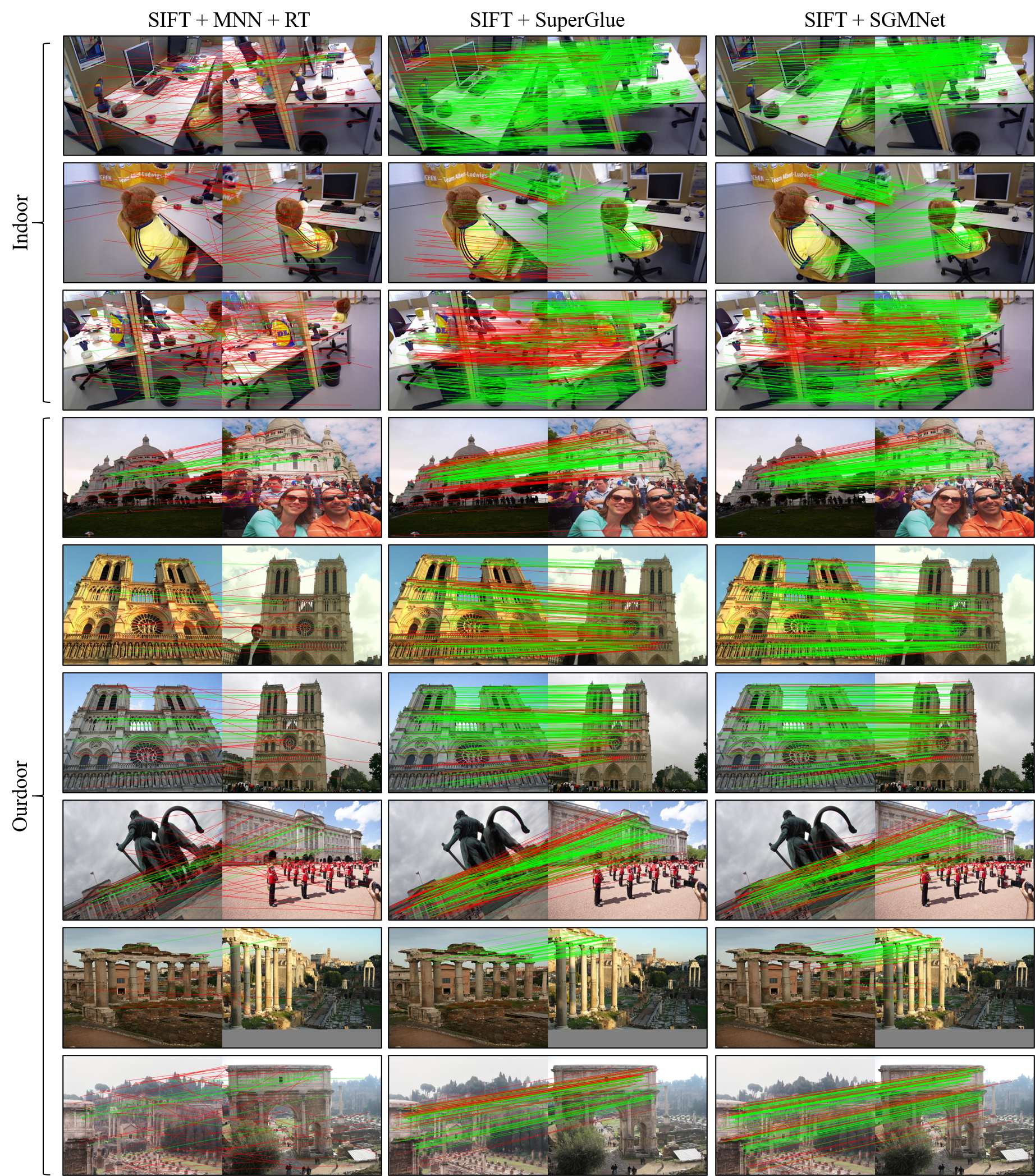}
	\caption{More visualizations.
	}
	\label{fig:more_vis}
\end{figure*}

{\small
\bibliographystyle{ieee_fullname}

}
\end{document}